\documentclass[10pt,twocolumn,letterpaper]{article}

\usepackage{iccv}              

\usepackage[symbol]{footmisc}
\usepackage{makecell}
\usepackage{multirow}
\usepackage{colortbl}
\usepackage{bm}
\usepackage{soul}
\usepackage{algorithm}
\usepackage{algpseudocode}
\usepackage{breakurl}

\definecolor{iccvblue}{rgb}{0.21,0.49,0.74}
\usepackage[pagebackref,breaklinks,colorlinks,allcolors=iccvblue]{hyperref}

\def\paperID{6703} 
\def\confName{ICCV}
\def\confYear{2025}

\title{Adversarial Distribution Matching for Diffusion Distillation \\Towards Efficient Image and Video Synthesis}

\author{
\textbf{
Yanzuo Lu\textsuperscript{\rm 1,2},
Yuxi Ren\textsuperscript{\rm 2},
Xin Xia\textsuperscript{\rm 2},
Shanchuan Lin\textsuperscript{\rm 2},
Xing Wang\textsuperscript{\rm 2},
} \\
\textbf{
Xuefeng Xiao\textsuperscript{\rm 2$*$},
Andy J. Ma\textsuperscript{\rm 1,3,4$\dagger$}, 
Xiaohua Xie\textsuperscript{\rm 1,3,4,5}, 
Jian-Huang Lai\textsuperscript{\rm 1,3,4,5}}\\\\
\textsuperscript{\rm 1}Sun Yat-Sen University \quad
\textsuperscript{\rm 2}ByteDance Seed Vision\\
\textsuperscript{\rm 3}Guangdong Provincial Key Laboratory of Information Security Technology, China\\
\textsuperscript{\rm 4}Key Laboratory of Machine Intelligence and Advanced Computing, Ministry of Education, China\\
\textsuperscript{\rm 5}Pazhou Lab (HuangPu), Guangzhou, China\\
{\tt\small oliveryanzuolu@gmail.com,xiaoxuefeng.ailab@bytedance.com,majh8@mail.sysu.edu.cn}
\vspace{-0.5cm}
}

\begin{document}

\setlength{\abovedisplayskip}{4pt}
\setlength{\belowdisplayskip}{4pt}

\maketitle

\footnotetext[1]{Project Lead.\quad$\dagger$ Corresponding Author.}

\begin{abstract}
Distribution Matching Distillation (DMD) is a promising score distillation technique that compresses pre-trained teacher diffusion models into efficient one-step or multi-step student generators.
Nevertheless, its reliance on the reverse Kullback-Leibler (KL) divergence minimization potentially induces mode collapse (or mode-seeking) in certain applications.
To circumvent this inherent drawback, we propose \textbf{Adversarial Distribution Matching (ADM)}, a novel framework that leverages diffusion-based discriminators to align the latent predictions between real and fake score estimators for score distillation in an adversarial manner.
In the context of extremely challenging one-step distillation, we further improve the pre-trained generator by adversarial distillation with hybrid discriminators in both latent and pixel spaces.
Different from the mean squared error used in DMD2 pre-training, our method incorporates the distributional loss on ODE pairs collected from the teacher model, and thus providing a better initialization for score distillation fine-tuning in the next stage.
By combining the adversarial distillation pre-training with ADM fine-tuning into a unified pipeline termed \textbf{DMDX}, our proposed method achieves superior one-step performance on SDXL compared to DMD2 while consuming less GPU time.
Additional experiments that apply multi-step ADM distillation on SD3-Medium, SD3.5-Large, and CogVideoX set a new benchmark towards efficient image and video synthesis.



\end{abstract} 
\vspace{-0.5cm}
\section{Introduction}
\label{sec:introduction}

\begin{figure}
    \centering
    \includegraphics[width=\linewidth]{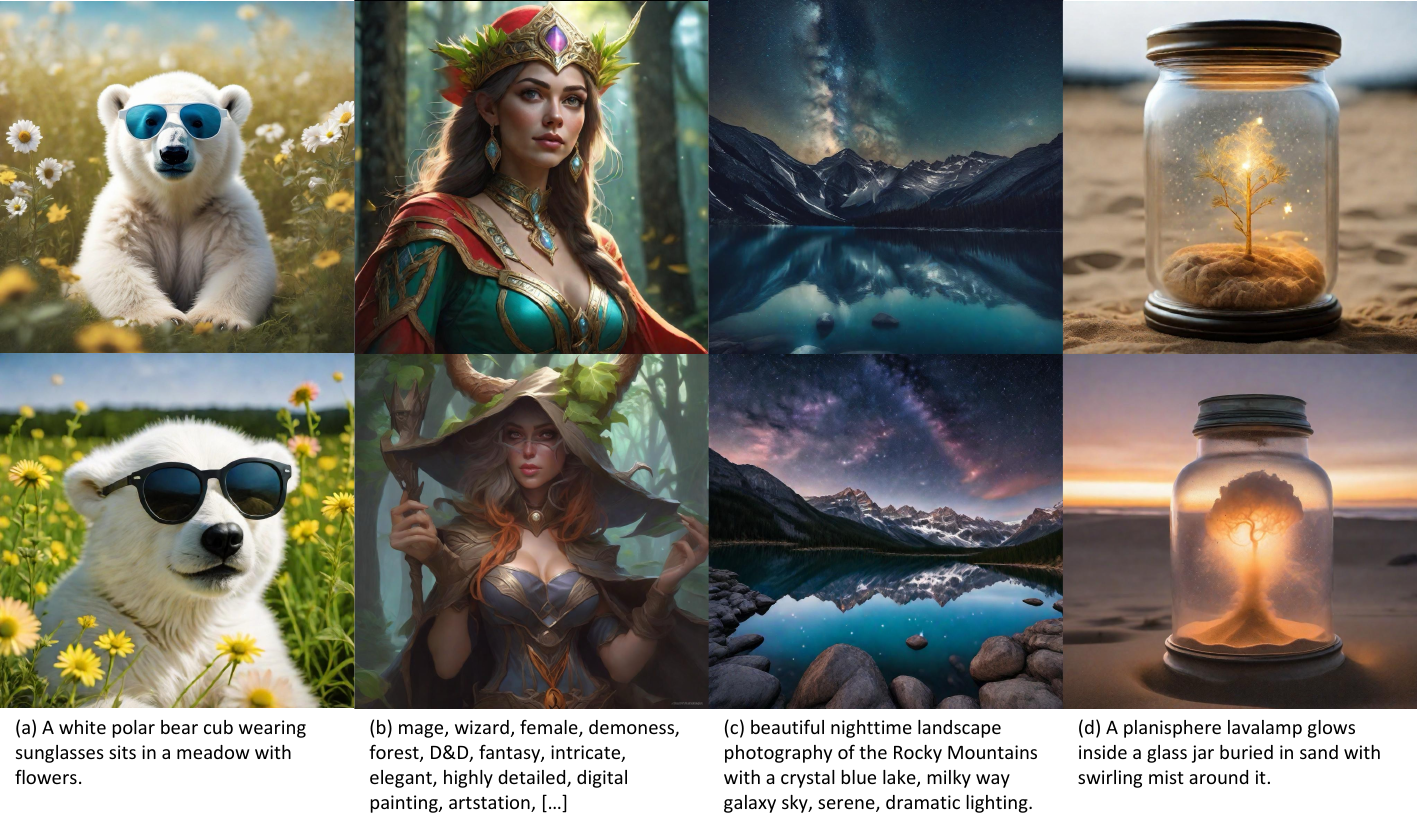}
    \vspace{-0.8cm}
    \caption{
        In these images, some are generated by \textbf{the baseline SDXL via 50NFE}, the others with \textbf{our DMDX in 1NFE}. Can you tell which is the accelerated one? Answers in the footnote\protect\hyperlink{footnote3}{\protect\footnotemark[3]}. 
    }
    \vspace{-0.5cm}
    \label{fig:teaser}
\end{figure}

Recent methods for accelerating diffusion models~\cite{ho2020denoising, karras2022elucidating, song2021denoising, song2021scorebased} have concentrated on reducing the sampling steps through distillation.
The distillation process generally trains a more efficient generator (a.k.a student model) that approximates the output distribution of the pretrained teacher.
This field has observed multiple pathways developing in parallel with progressive distillation~\cite{lin2024sdxllightning,salimans2022progressive}, consistency distillation~\cite{kim2024consistency,luo2023latent,luo2023lcmlora,song2023consistency,wang2024phased,zheng2024trajectory,mao2024osv,wang2024animatelcm,ren2024hypersd}, score distillation~\cite{chadebec2024flash,kohler2024imagine,sauer2023adversarial,yin2024improved,yin2024onestep,zhou2024score,luo2024onestep,jayashankar2025scoreofmixture,yi2025magic,yin2024slow,salimans2024multistep,ren2024hypersd}, rectified flow~\cite{liu2022flow,liu2024instaflow,lu2024simplifying,wang2024rectified,yan2024perflow} and adversarial distillation~\cite{lin2024sdxllightning,ren2024hypersd,sauer2023adversarial,sauer2024fast,yin2024improved,mao2024osv,zhang2024sfv,lin2025diffusion}.
They are currently appearing as distinct yet non-exclusive research directions.

\footnotetext[2]{\protect\hypertarget{footnote3}{}\textbf{Our DMDX (left to right):} bottom, top, bottom, top.}

{Distribution Matching Distillation (DMD)}~\cite{yin2024onestep,yin2024improved} is a promising score distillation~\cite{poole2023dreamfusion} approach, which distills the powerful text-to-image diffusion model SDXL-Base~\cite{rombach2022highresolution} into a one-step generator with great fidelity.
While DMD's primary contribution lies in introducing additional regularizers to constrain the distribution matching loss, its predecessor Diff-Instruct~\cite{luo2023diff} pioneered the use of a fake score estimator to approximate the student model's output distribution. 
This contrasts with the distillation loss in Adversarial Diffusion Distillation (ADD)~\cite{sauer2023adversarial}, which directly employs the student model itself as score estimator.
Theoretically, an intrinsic correspondence for DMD and the distillation loss in ADD can be equivalently found between {Variational Score Distillation (VSD)}~\cite{wang2023prolificdreamer} and {Score Distillation Sampling (SDS)}~\cite{poole2023dreamfusion} in text-to-3D generation, where SDS represents a specialized instance of VSD through the use of a single-point Dirac distribution as the variational distribution~\cite{wang2023prolificdreamer}.
And thus the VSD loss in DMD performs better than the SDS loss in ADD.
Intuitively, since the capacity of a diffusion model decreases significantly when distilled for few-step generations~\cite{lin2024sdxllightning}, the student model no longer serves as a decent score estimator as the teacher.
Therefore, the score-based distillation loss in ADD hardly contributes to its final performance.

However, the optimization of DMD loss relies on the reverse {Kullback-Leibler (KL)} divergence minimization which is zero-forcing that drives low-probability regions to zero, causing the model to focus on only a few dominant modes and potentially leading to mode collapse~\cite{minka2005divergence}.
To enhance sample diversity, DMD~\cite{yin2024onestep} employs an {ODE-based (Ordinary Differential Equation)} regularizer with synthetic data, while DMD2~\cite{yin2024improved} introduces a {GAN-based (Generative Adversarial Network~\cite{goodfellow2014generative})} regularizer with real data to counterbalance this side effect.
Subsequent efforts such as {Moment Matching Distillation (MMD)}~\cite{salimans2024multistep}, {Score identity Distillation (SiD)}~\cite{zhou2024score} and {Score Implicit Matching (SIM)}~\cite{luo2024onestep} all used a variant of Fisher divergence to align the fake score estimator with pre-trained real score estimator~\cite{jayashankar2025scoreofmixture}.
Despite great success, their ability for distribution matching is limited by their dependence on a predefined form of explicit divergence metric.
In this work, we would like to raise and investigate a question: \ul{Can we bypass the limitations of a predefined divergence by developing a framework that learns an implicit, data-driven discrepancy measure, thereby enabling more flexible and fine-grained matching of complex, high-dimensional distributions?}

Since the multifaceted alignment requirements in complex multimodal text-conditioned image or even video generation may not be fully captured, this motivates our exploration of more adaptive discrepancy learning paradigms.
As our \ul{first contribution}, we show how to align the latent predictions between real and fake score estimators to realize score distillation in an adversarial manner with prior knowledge of teacher model and dynamically learnable parameters, termed \textbf{Adversarial Distribution Matching (ADM)}.
This is different from the ODE-based or GAN-based regularizer in DMD and DMD2 that is used to counterbalance the mode collapse effect of reverse KL divergence.
In contrast, we are performing distribution matching by means of GAN training to replace and circumvent the use of DMD loss, which we will provide more discussion in \cref{sec:relation_dmd_dmd2}.

Our \ul{second contribution} concerns about the extremely challenging one-step distillation, where we observe that the score distillation has a higher risk of gradient exploding and vanishing.
We attribute the issue more to the lower overlap of support sets between the student and teacher distributions, not just the approximation errors in the fake score estimator as attributed in DMD2~\cite{yin2024improved}.
In other words, while score distillation yields superior generation quality, it imposes higher requirements on initialization especially when distilled in extremely few step, which we will provide more in-depth analysis in \cref{sec:importance_pretrain}.
Although we notice that an ODE-based pre-training on synthetic data is used for SDXL one-step distillation in DMD2  \href{https://github.com/tianweiy/DMD2/blob/main/experiments/sdxl/README.md#1-step-sample-trainingtesting-commands-work-in-progress:~:text=For%201%2Dstep%20model%2C%20we%20need%20an%20extra%20regression%20loss%20pretraining.}{implementation}, the mean squared error loss was probably still not enough to provide more overlapping regions of support sets between the student and teacher distributions.
Our experiments demonstrate that when providing a better initialization for distribution matching, the effect of Two Time-scale Update Rule (TTUR)~\cite{bynagari2017gans} to the final performance is very limited.

Naturally, our \ul{third contribution} focuses on providing a better initialization for further score-based fine-tuning.
We employ adversarial distillation to trade off sample quality and mode converge inspired by SDXL-Lightning~\cite{lin2024sdxllightning} and LADD~\cite{sauer2024fast}.
With this distribution-level loss optimization, we can pre-train the student model to capture more potential modes of the teacher model distribution, especially through a hybrid discriminator in both latent and pixel spaces which we will introduce later in \cref{sec:ADP}.
To facilitate the diversity, we also propose employing a cubic timestep schedule for the generator to bias towards higher noise levels.


By combining adversarial distillation pre-training with ADM fine-tuning into a unified pipeline termed \textbf{DMDX}, our one-step SDXL provides competitive fidelity compared to the baseline with \textbf{50$\times$ acceleration} in the Number of Function Evaluation (NFE) as shown in \cref{fig:teaser}.
More experiments on multi-step ADM distillation across the best existing diffusion models including SD3-Medium, SD3.5-Large~\cite{esser2024scaling}, and CogVideoX~\cite{yang2024cogvideox} consistently achieve new benchmarks for efficient image and video synthesis.

\section{Related Work}
\label{sec:related_work}

\noindent\textbf{Progressive Distillation} was proposed in \cite{salimans2022progressive} to distill multi-step prediction into one-step prediction of the same distance along the trajectory. 
SDXL-Lightning~\cite{lin2024sdxllightning} extended this idea by utilizing GAN training and achieved one-step high-resolution (1024px) generation for the first time.
This approach involving multi-stage process can be quite cumbersome, as it requires iteratively distilling from its predecessor, halving the sampling steps each time.


\noindent\textbf{Consistency Distillation} proposed~\cite{song2023consistency,geng2024consistency,song2023improved,luo2023latent,luo2023lcmlora} to enforce the consistency property into diffusion model, i.e. predictions towards original sample are consistent for arbitrary pairs of noisy timesteps belong to the same trajectory. 
Subsequent efforts extended it into the trajectory consistency to relax the training objectives, i.e. predictions towards noisy samples of arbitrary subsequent timesteps are consistent, including CTM~\cite{kim2024consistency}, TCD~\cite{zheng2024trajectory}, TSCD~\cite{ren2024hypersd} and PCM~\cite{wang2024phased}.


\noindent\textbf{Rectified Flow}~\cite{liu2022flow,liu2024instaflow} aims to obtain faster straight trajectories through multiple reflow processes that iteratively learn the velocity of many ODE pairs from its predecessor.
PeRFlow~\cite{yan2024perflow} tried splitting the trajectory into pieces and applying piece-wise rectification with real data samples.
In this paper, we empirically found the straightness can also be satisfied via an adversarial distillation paradigm without the need for repeatedly collecting tremendous synthetic data.

\noindent\textbf{Score Distillation} 
intuitively tries to keep that a sample appearing in the student model distribution at a specific noise level is with the same probability as it does in the teacher model distribution,.
In light of the motivation that relying on a single divergence might be problematic, a concurrent work Score-of-Mixture Distillation (SMD)~\cite{jayashankar2025scoreofmixture} shares our perspective and explicitly designed a class of $\alpha$-skew Jensen–Shannon (JS) divergences for optimization.
In contrast, we implicitly measure the discrepancy between the fake and real score estimators in an adversarial manner, facilitating a more capable and adaptive discrepancy learning paradigm along the distillation process.

\noindent\textbf{Adversarial Distillation} employs a discriminator to align the student model distribution with a specific target distribution~\cite{luo2024you,zhou2024adversarial}. 
While methods UFOGen~\cite{xu2024ufogen}, DMD2~\cite{yin2024improved}, and APT~\cite{lin2025diffusion} directly align with the real data distribution, SDXL-Lightning~\cite{lin2024sdxllightning} and Hyper-SD~\cite{ren2024hypersd} utilize intermediate timestep predictions from the teacher model as approximation objectives.
Inspired by LADD~\cite{sauer2024fast}, our adversarial distillation pre-training aligns with ODE-based synthetic data generated from the teacher model.

\noindent\textbf{Human Feedback Learning} was firstly considered by ReFL~\cite{xu2023imagereward} in diffusion models that includes two stages of reward model training and preference fine-tuning. 
Hyper-SD~\cite{ren2024hypersd} treated this as a standalone technique, ultimately applying LoRA insertion to shift the output distribution of the low-step generator toward human preferences.
Subsequent studies, notably \cite{luo2024diff,luo2024david}, further attempted to integrate CFG and Score-based divergence.

\section{Preliminaries}
\label{sec:preliminaries}

\subsection{Diffusion Model} 

Given a training data distribution $p_{\text{data}}$ with standard deviation $\sigma_{\text{data}}$, the diffusion model~\cite{ho2020denoising} generates samples by reversing the forward diffusion process that progressively adds noise to a data sample $\bm{x}_0\sim p_{data}$ as, 
\begin{equation}\label{eq:diffusion}
    \bm{q}(\bm{x}_t|\bm{x}) \sim \mathcal{N}(\alpha_t \bm{x}, \sigma_t^2 \bm{I}), 
\end{equation}
where $\alpha_t\ge 0,\sigma_t>0$ are specified noise schedules such that $\alpha_t/\sigma_t$ satisfies monotonically decreasing w.r.t. $t$ and larger $t$ indicates greater noise.
We consider two different formulations for denoising models.

\noindent\textbf{DDPM and DDIM}~\cite{ho2020denoising,song2021denoising} assume discrete-time schedules with $t\in[1,T]$ (typically $T=1000$) and noise prediction parameterization\footnote{This is not mandatory for works like CogVideoX~\cite{yang2024cogvideox} which also used DDIM sampling but parameterized in velocity. 
The different parameterizations can be equivalently transformed into each other~\cite{salimans2022progressive}.
}~\cite{salimans2022progressive}. 
The training objective is given by,
\begin{equation}\label{eq:ddim_pretrain_loss}
    \mathbb{E}_{\bm{x}_0,t,\epsilon\sim\mathcal{N}(\bm{0},\bm{I})}[w(t)\|\bm{\epsilon}_\theta(\bm{x}_t,t)-\epsilon\|_2^2],
\end{equation}
where $w(t)$ is a weighting function and $\bm{\epsilon}_\theta$ is a neural network with parameters $\theta$. 
The noise schedule is defined as $\alpha_t=\sqrt{\bar\alpha_t}, \sigma_t=\sqrt{1-\bar\alpha_t}$ such that $\bm{x}_t=\sqrt{\bar\alpha_t}\bm{x}_0+\sqrt{1-\bar\alpha_t}\epsilon$.
For DDIM sampling, it solves the \textit{Probability-Flow Ordinary Differential Equations (PF-ODE)}~\cite{song2021scorebased} by $d\bar{\bm{x}}_t=\bm{\epsilon}_\theta(\frac{\bar{\bm{x}}_t}{\sqrt{\bar{\sigma}_t^2+1}})d\bar{\sigma}_t$, where $\bar{\bm{x}}_t=\frac{\bm{x}_t}{\sqrt{\bar\alpha_t}}$ and $\bar\sigma_t=\sqrt{\frac{1-\bar\alpha_t}{\bar\alpha_t}}$, starting from $\bm{x}_T\sim\mathcal{N}(\bm{0}, \bm{I})$ and stopping at $\bm{x}_0$.

\noindent\textbf{Flow Matching}~\cite{lipman2023flow,liu2022flow,lu2024simplifying} uses velocity prediction parameterization~\cite{salimans2022progressive} and continuous-time coefficients (typically $\alpha_t=1-t, \sigma_t=t$ with $t\in[0,T=1]$).
The conditional probability path or the velocity we say can be given as $\bm{v}_t=\frac{d\alpha_t}{dt}\bm{x}_0+\frac{d\sigma_t}{dt}\epsilon$, such that
the training objective is,
\begin{equation}\label{eq:flowmatch_pretrain_loss}
    \mathbb{E}_{\bm{x}_0,t,\epsilon\sim\mathcal{N}(\bm{0},\bm{I})}[w(t)\|\bm{v}_\theta(\bm{x}_t,t)-\bm{v}_t\|_2^2],
\end{equation}
where $w(t)$ is a weighting function and $\bm{v}_\theta$ is a neural network parameterized by $\theta$.
The sampling procedure starts from $t=T$ with $\bm{x}_T\sim\mathcal{N}(\bm{0}, \bm{I})$ and stops at $t=0$, solving the PF-ODE by $d\bm{x}_t=\bm{v}_\theta(\bm{x}_t,t)dt$.

\begin{figure*}[t]
    \centering
    \includegraphics[width=0.875\linewidth]{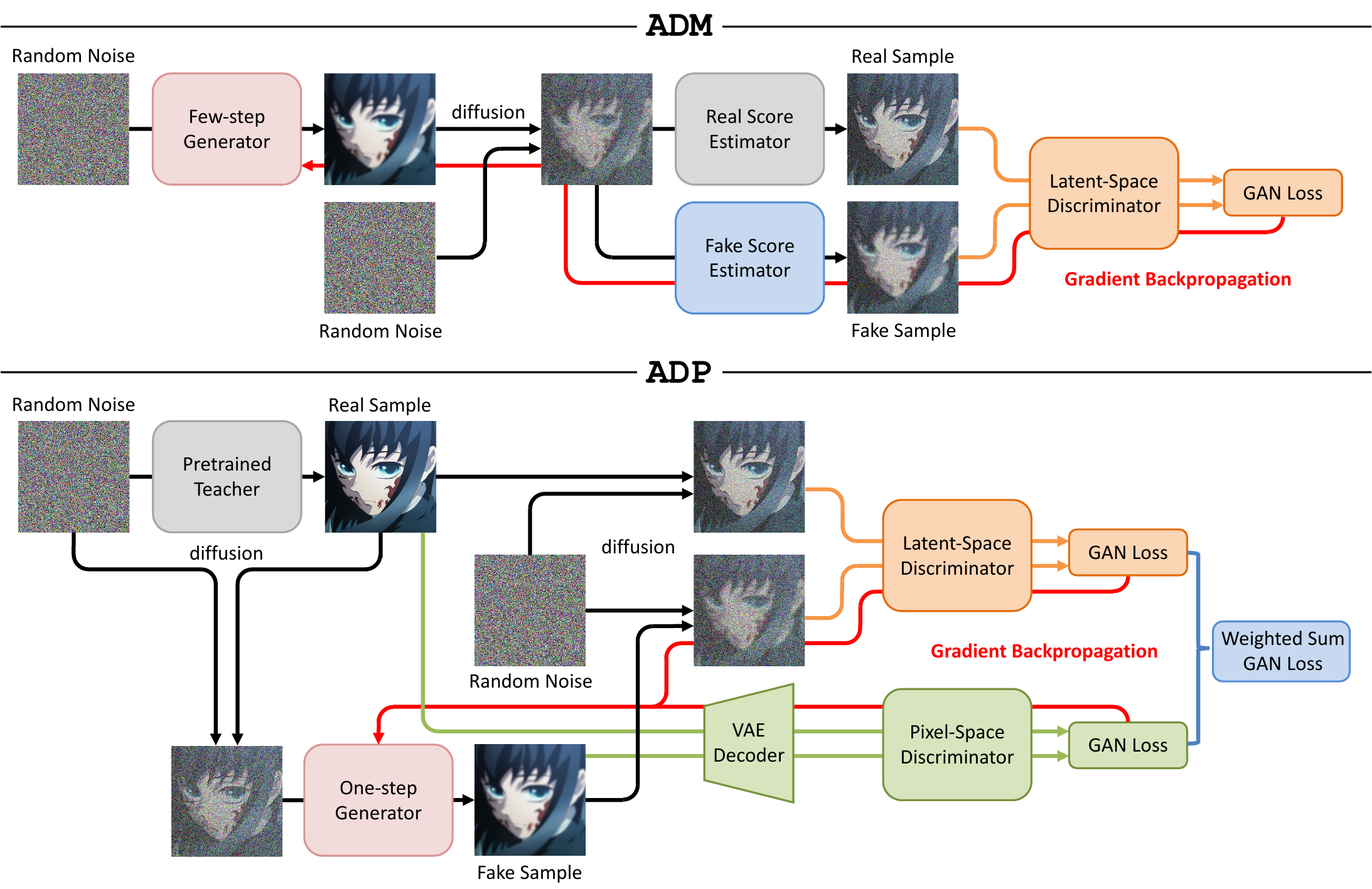}
    \vspace{-0.3cm}
    \caption{Overall pipeline of our proposed Adversarial Distribution Matching (\textbf{ADM}) and Adversarial Distillation Pre-training (\textbf{ADP}).}
    \vspace{-0.4cm}
    \label{fig:pipeline}
\end{figure*}

\subsection{Distribution Matching Distillation}
\label{sec:dmd}

DMD~\cite{yin2024onestep,yin2024improved} distills pretrained diffusion models $\bm{F}_\phi(\bm{x}_t,t)$ into one-step or multi-step efficient generators $\bm{G}_\theta(\bm{x}_t,t)$ by minimizing the reverse KL divergence between the target distribution $p_{\text{real}}$ and the efficient generator output distribution $p_{\text{fake}}$.
The gradient of DMD objective w.r.t. $\theta$ is,
\begin{equation}\resizebox{0.9\hsize}{!}{$
\label{eq:dmd_gradient}
    \nabla_\theta \mathcal{L}_{\text{DMD}} = \mathop\mathbb{E}\limits_{\bm{z},t^\prime,t,\bm{x}_{t}}-[(s_{\text{real}}(\bm{x}_{t})-s_{\text{fake}}(\bm{x}_{t})) \frac{d\bm{G}_\theta(\bm{z},t^\prime)}{d\theta}],
$}\end{equation}
where $\bm{z}\sim\mathcal{N}(\bm{0},\bm{I})$, $t^\prime$ is randomly selected from predefined generator schedule, $t\sim \mathcal{U}(0,T)$, and noisy samples $\bm{x}_{t}=\bm{q}(\bm{x}_{t}|\hat{\bm{x}}_0)$ are obtained by randomly diffusing the generator output $\hat{\bm{x}}_0=\bm{G}_\theta(\bm{z}, t^\prime)$.
Score functions $s_{\text{real}}(\bm{x}_{t})=\nabla_{\bm{x}_{t}}\log p_{\text{real}}(\bm{x}_{t})$, $s_{\text{fake}}(\bm{x}_{t})=\nabla_{\bm{x}_{t}}\log p_{\text{fake}}(\bm{x}_{t})$ are vector fields that point towards higher density of data at a given noise level~\cite{karras2022elucidating,song2021scorebased} for $p_{\text{real}}$ and $p_{\text{fake}}$, respectively.

While the real score estimator is the teacher model $\bm{F}_\phi(\bm{x}_t,t)$ itself, the fake score estimator $\bm{f}_\psi(\bm{x}_t,t)$ is initialized the same as $\bm{F}_\phi(\bm{x}_t,t)$ and dynamically learned to describe $p_{\text{fake}}$ with pretrain loss as \cref{eq:ddim_pretrain_loss,eq:flowmatch_pretrain_loss}.
In practice, the gradient in \cref{eq:dmd_gradient} is computed as,
\begin{equation}\label{eq:dmd_gradient_imple}
    grad(\hat{\bm{x}}_0,\bm{x}_{t},{t})=\frac{\bm{f}_\psi(\bm{x}_{t},{t})-\bm{F}_\phi(\bm{x}_{t},{t})}{\|\hat{\bm{x}}_0-\bm{F}_\phi(\bm{x}_{t},{t})\|_1}
\end{equation}
such that the training loss is implemented like,
\begin{equation}\resizebox{0.9\hsize}{!}{$\label{eq:dmd_loss}
    \mathcal{L}_{\text{DMD}}(\theta)=\mathop\mathbb{E}\limits_{\bm{z},t^\prime,t,\bm{x}_t}[\|\hat{\bm{x}}_0-sg(\hat{\bm{x}}_0-grad(\hat{\bm{x}}_0,\bm{x}_{t},{t}))\|_2^2],
$}\end{equation}
where $sg(\cdot)$ denotes the stop gradient operation. 


\section{Methodology}
\label{sec:methodology}

\subsection{Adversarial Distribution Matching}

Instead of using a predefined divergence between the fake and real distributions, we use an implicit, data-driven discrepancy measure through an adversarial discriminator.
Specifically, our discriminator $\bm{D}_\tau(\bm{x}_t,t)$ consists of a frozen latent diffusion model initialized the same as teacher model $\bm{F}_\phi(\bm{x}_t,t)$ and multiple trainable heads added upon different UNet~\cite{ronneberger2015unet} or DiT~\cite{peebles2023scalable} blocks.
Given the noisy sample $\bm{x}_{t}=\bm{q}(\bm{x}_{t}|\hat{\bm{x}}_0)$ diffused from the output of few-step generator $\hat{\bm{x}}_0=\bm{G}_\theta(\bm{z}, t^\prime)$, the score estimators no longer solve the PF-ODE w.r.t. the end point ${\bm{x}}_0^{\text{fake}}=\bm{f}_\psi(\bm{x}_{t},{t})$ and ${\bm{x}}_0^{\text{real}}=\bm{F}_\phi(\bm{x}_{t},{t})$ as used in \cref{eq:dmd_gradient_imple}.

Instead, we set a fixed timestep interval $\Delta t$ (defaults to $T/64$) and solve the PF-ODE w.r.t. $(t-\Delta t)$, such that the fake sample ${\bm{x}}_{t-\Delta t}^{\text{fake}}$ and real sample ${\bm{x}}_{t-\Delta t}^{\text{real}}$ can be obtained to serve as score predictions and sent into the discriminator.
The discriminator hierarchically aggregates features from frozen backbone layers and dynamically weights them through multiple learnable heads, establishing an adaptive discrepancy metric that leverages both diffusion priors and data-driven trainable dynamics.
We use Hinge loss~\cite{lim2017geometric} to train the generator $\bm{G}_\theta(\bm{x}_t,t)$ and discriminator $\bm{D}_\tau(\bm{x}_t,t)$ alternatively.
This encourages the fake score prediction $\bm{x}_{t-\Delta t}^{\text{fake}}$ to be closer to the real score prediction $\bm{x}_{t-\Delta t}^{\text{real}}$:
\begin{equation}\resizebox{0.6\hsize}{!}{$
    \mathcal{L}_{\text{GAN}}(\theta)=\mathop\mathbb{E}\limits_{\bm{x}_{t-\Delta t}^{\text{fake}}}[-\bm{D}_\tau({\bm{x}}_{t-\Delta t}^{\text{fake}}, t-\Delta t)]\label{eq:gan_generator_loss}
$}\end{equation}
\begin{equation}\resizebox{0.85\hsize}{!}{$
    \begin{aligned}\label{eq:gan_discriminator_loss}
        \mathcal{L}_{\text{GAN}}(\tau)=\mathop\mathbb{E}\limits_{\bm{x}_{t-\Delta t}^{\text{fake}},\bm{x}_{t-\Delta t}^{\text{real}}}[&\max (0, 1+\bm{D}_\tau({\bm{x}}_{t-\Delta t}^{\text{fake}}, t-\Delta t))\\
        +&\max(0,1-\bm{D}_\tau({\bm{x}}_{t-\Delta t}^{\text{real}}, t-\Delta t))]
    \end{aligned}
$}\end{equation}
In conjunction with the dynamically learned fake model, we clarify our training procedure in \cref{app:algorithm}. The overall pipeline is demonstrated in  \cref{fig:pipeline}.


\subsubsection{Motivation of Discriminator Timestep \texorpdfstring{$(t-\Delta t)$}{(t-Delta t)}}
\label{sec:adm_timestep}
Given that the final objective of score distillation is to match the probability flows varying with noise levels of student and teacher models to exactly the same, timestep information must be considered when measuring the discrepancy between distributions.
This coincides with our discriminator design that uses a pre-trained diffusion model, and we take a small step alongside the PF-ODE, succeeding in preserving the input timestep information of score estimators.

\subsubsection{Data-driven Effect}
\label{sec:adm_data}
The flexibility of using discriminators for distributional discrepancy measure is not only in the noise level of score function, but also within the distillation process.
As distillation iterates, the model is exposed to increasingly diverse data, causing the discrepancies in modes between the two distributions to change.
In the early phase of training where the discrepancy is significant, a more global evaluation is necessary, whereas in the later phases, when the discrepancy becomes minor, a more localized, fine-grained optimization might become essential.
In other words, driven by the volume of data, the divergence measure employed across different training phases can vary.

\subsubsection{Relationship with DMD and DMD2}
\label{sec:relation_dmd_dmd2}

To mitigate the mode collapse issue in DMD loss, an ODE-based regularizer and a GAN-based regularizer are additionally used for distillation in DMD~\cite{yin2024improved} and DMD2~\cite{yin2024improved}, respectively.
However, these two regularizers do not fundamentally address the mode-seeking behavior introduced by reverse KL divergence as shown in \cref{fig:kl}(a), but rather counterbalance it by trade-offs between losses.
In ADM, our adversarial loss is actually playing the role of DMD loss to realize score distillation with an implicit, data-driven discrepancy measure instead of a predefined divergence.
Therefore, our motivation for using GAN training in ADM is different from that of DMD2~\cite{yin2024improved} and we don't require additional regularizers.

\begin{figure}
    \centering
    \includegraphics[width=\linewidth]{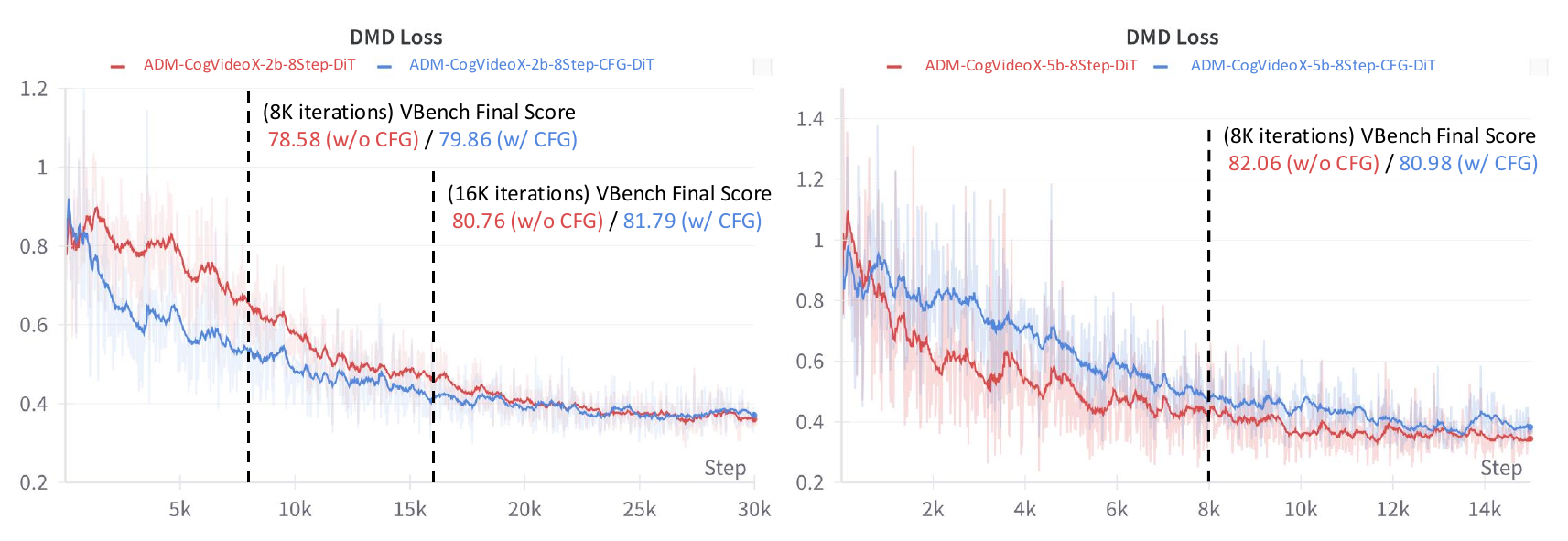}
    \vspace{-0.8cm}
    \caption{
        Changes of DMD loss over multi-step ADM distillation for CogVideoX.
        Note that we did not optimize this objective directly during ADM distillation but recorded it over iterations.
    }
    \vspace{-0.5cm}
    \label{fig:ablation_relation_dmd_dmdx}
\end{figure}

Intuitively, the learnable discriminator can approximate any nonlinear function to implicitly measure distribution divergence, which probably inherently encompasses the reverse KL divergence in DMD loss.
As shown in \cref{fig:ablation_relation_dmd_dmdx}, we visualize the changes of DMD loss in \cref{eq:dmd_loss} during the multi-step ADM distillation on CogVideoX~\cite{yang2024cogvideox}.
Though not directly optimize on \cref{eq:dmd_loss}, the results indicate a very steady downward trend that supports our assumption.
More theoretical discussion is provided in \cref{sec:discussion_discrepancy}.

\subsection{Adversarial Distillation Pre-training}
\label{sec:ADP}



To stabilize the extremely difficult one-step distillation, we opt to provide a better initialization for ADM fine-tuning with adversarial distillation pre-training on synthetic data.
Our pre-training configuration refer to Rectified Flow~\cite{liu2022flow} in several aspects, where we
\textbf{1)} collect the ODE pairs from teacher model in an offline manner,
\textbf{2)} construct noisy samples by linearly interpolation between pure noise and clean data sample of the ODE pair,
\textbf{3)} alter the prediction target of the generator to the velocity of ODE pair.

As for adversarial training, we formulate a latent-space discriminator $\bm{D}_{\tau_1}(\bm{x}_t,t)$ initialized from teacher model and a pixel-space discriminator $\bm{D}_{\tau_2}(\bm{x})$ initialized from the vision encoder of SAM~\cite{kirillov2023segment} model, respectively, as shown in \cref{fig:pipeline}.
We also append multiple trainable heads to both the backbone networks similar to the practice in ADM.
All these contribute to increasing the discriminative capability, facilitating the student model in discovering more potential modes in the teacher model distribution.
Specifically, let $\tilde{\bm{x}}_0=\bm{G}_\theta(\bm{x}_t,t)$ denote the predicted PF-ODE endpoint of generator, where $\bm{x}_t$ represents the noisy sample interpolated between an ODE pair $(\bm{x}_T,\bm{x}_0)$ with random timestep $t\in[0,T]$.
For latent-space discriminator, we diffuse the generator output with another random noise and timestep $t^\prime\in(0,T]$, getting $\tilde{\bm{x}}_{t^\prime}=\bm{q}(\tilde{\bm{x}}_{t^\prime}|\tilde{\bm{x}}_0)$ as its input.
For pixel-space discriminator, the generator output will be first decoded via VAE decoder and then sent into the vision encoder.
The training objective, also based on Hinge loss~\cite{lim2017geometric}, encourages the generator output $\tilde{\bm{x}}_0$ to be closer to the synthetic data sample $\bm{x}_0$:
\begin{equation}\resizebox{0.7\hsize}{!}{$
    \begin{aligned}
    \mathcal{L}_{\text{GAN}}(\theta)=\ \mathop\mathbb{E}\limits_{\tilde{\bm{x}}_0,t^\prime}-[&\lambda_1\bm{D}_{\tau_1}(\tilde{\bm{x}}_{t^\prime},t^\prime)+\lambda_2\bm{D}_{\tau_2}(\tilde{\bm{x}}_0)]
    \end{aligned}
$}\end{equation}
\begin{equation}\resizebox{0.75\hsize}{!}{$
    \begin{aligned}
        \mathcal{L}_{\text{GAN}}(\tau_1,\tau_2)=\mathop\mathbb{E}\limits_{\bm{x}_0,\tilde{\bm{x}}_0,t^\prime}[&\lambda_1\cdot\max(0,1+\bm{D}_{\tau_1}(\tilde{\bm{x}}_{t^\prime},t^\prime))\\
        +&\lambda_2\cdot\max(0,1+\bm{D}_{\tau_2}(\tilde{\bm{x}}_0))\\
        +&\lambda_1\cdot\max(0,1-\bm{D}_{\tau_1}(\bm{x}_{t^\prime},t^\prime))\\
        +&\lambda_2\cdot\max(0,1-\bm{D}_{\tau_2}(\bm{x}_0))]
    \end{aligned}
$}\end{equation}
We empirically found that setting balancing coefficients $\lambda_1=0.85,\lambda_2=0.15$ produces visually coherent results.

\subsubsection{Cubic Generator Timestep Schedule}
Intuitively, higher noise levels encourage exploration of new modes by weakening the restrictive information encoded in latent representations.
Therefore, we propose employing a cubic timestep schedule for the generator.
The schedule maps uniform $[0,T)$ samples through $[1-(t/T)^3]*T$, non-linearly concentrating values near $T$ with heavy noise similar to LADD~\cite{sauer2024fast}.

\subsubsection{Uniform Discriminator Timestep Schedule}
Unlike in ADM where the discriminator is firstly used for score distillation, the utilization of latent-space discriminator is common for adversarial distillation works.
Inspired by SDXL-Lightning~\cite{lin2024sdxllightning}, where they found that the diffusion encoder is trained to focus on high-frequency details at lower timesteps and low-frequency structures at higher timesteps, we set a uniform $(0,T]$ for discriminator timestep $t^\prime$ to capture both advantages during pre-training.

\subsubsection{Relationship with LADD}
Our motivation to perform adversarial distillation on synthetic data is inspired by LADD~\cite{sauer2024fast}, but differs in a lot that we
\textbf{1)} construct noisy samples via ODE pairs in the style of Rectified Flow~\cite{liu2022flow} rather than random noise,
\textbf{2)} develop a cubic generator timestep schedule that facilitate deterministic Euler sampling rather than consistency ones,
\textbf{3)} introduce an additional pixel-space encoder to increase the capability of discriminator and find more modes.

\subsection{Discussion}
\label{sec:theoretical_analysis}

\subsubsection{Difference between ADM and ADP}


One question may appear like \ul{what is the difference between these two adversarial approaches with latent-space discriminators?}
The effectiveness in score distillation correlates with the fact that score function $\nabla_{\bm{x}}\log p(\bm{x};\sigma(t))$ is defined at different noise levels $\sigma(t)$.
In contrast, adversarial distillation only aligns the distribution of clean data samples when $t=0$.
While the latent-space discriminator in pre-training is capturing information at different scales and details by randomly diffusing the generator output, the crux for ADM is more than that by solving the PF-ODE of both score estimators.
In other words, ADM additionally supervises the complete denoising process through noisy samples that are with higher density at different noise levels of respective distribution.

This leads to a situation in ADM when two distributions initialized with less overlap in support sets, the noisy samples remain in regions unfamiliar to each other, and the discriminator can easily distinguish between them thus leading to extreme gradient signals.
However, since Gaussian noise is isotropic, we artificially create overlapping regions for the randomly diffused samples in ADP to make the discrimination more difficult, resulting in relatively smooth gradients.
Therefore, our ADM still falls under score distillation given that it encourages the entire probability flows to be closer, while the pre-training belongs to adversarial distillation because it only cares about the clean data distribution at $t=0$.

\subsubsection{Importance of Pre-training}
\label{sec:importance_pretrain}

Another question we haven't discussed so far is \ul{why do we need pre-training for one-step score distillation?}
Taking the reverse KL divergence used by DMD loss as an example:
\begin{equation}
    \mathbb{D}_{\text{KL}}(p_{\text{fake}}\|p_{\text{real}})=\int p_{\text{fake}}(\bm{x})\log \frac{p_{\text{fake}}(\bm{x})}{p_{\text{real}}(\bm{x})}d\bm{x}.
\end{equation}
When employing one-step distillation, the generator output is worse in visual fidelity and structural integrity compared to multi-step sampling, leading $p_{\text{fake}}(\bm{x})\rightarrow 0$ where $p_{\text{real}}(\bm{x})>0$ in regions.
The integrand $p_{\text{fake}}(\bm{x})\log \frac{p_{\text{fake}}(\bm{x})}{p_{\text{real}}(\bm{x})}$ approaching zero $0\cdot (-\infty)$ causes the optimization to avoid regions where $p_{\text{real}}(\bm{x})>0$ but $p_{\text{fake}}(\bm{x})$ has negligible density, a phenomenon called \textbf{zero-forcing}.
Instead of fully covering $p_{\text{real}}$'s support, $p_{\text{fake}}$ collapses to a subset of modes of $p_{\text{real}}$, inducing \textbf{mode-seeking} behavior as illustrated in \cref{fig:kl}(a).
During training this manifests itself as gradient vanishing sometimes.
And conversely, the fuzzy samples that the diffusion model typically produces through one step are also not within the teacher model distribution, yielding $p_{\text{real}}(\bm{x})\rightarrow 0$ where $p_{\text{fake}}(\bm{x})>0$ in regions and the integrand $p_{\text{fake}}(\bm{x})\log \frac{p_{\text{fake}}(\bm{x})}{0}$ diverging to $+\infty$, and thus resulting in numerical instability and gradient exploding. 

Similarly, when the support sets of the student and teacher distributions overlap hardly at all, forward KL divergence approaches $+\infty$ where $p_{\text{fake}}(\bm{x})>0$, while JS divergence saturates to a constant $\log 2$ and Fisher divergence may degenerate without definition.
Therefore, when this assumption is undermined, many of the single divergences no longer apply and a better initialization with more overlapping regions becomes essential as exemplified in \cref{fig:kl}(b).

\subsubsection{Theoretical Objective}\label{sec:discussion_discrepancy}
\begin{figure}
    \centering
    \includegraphics[width=\linewidth]{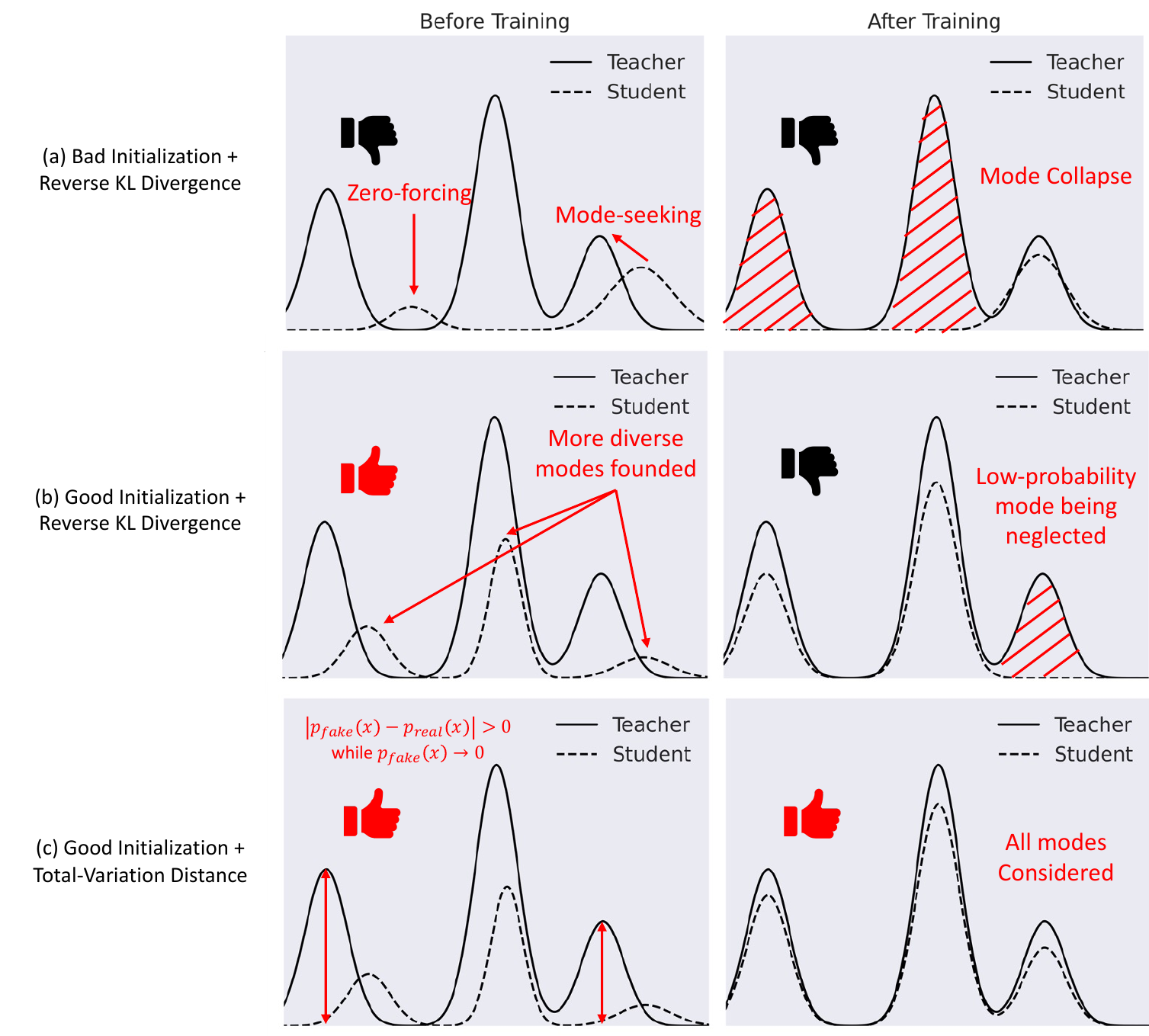}
    \vspace{-0.7cm}
    \caption{Illustration for theoretical discussion.}
    \vspace{-0.35cm}
    \label{fig:kl}
\end{figure}

The last question is \ul{why ADM is better than DMD loss theoretically?}
In fact, the Hinge GAN~\cite{lim2017geometric} we use has been proven to minimize Total-Variation Distance (TVD)~\cite{tan2019calibrated}:
\begin{equation}
    TV(p_{\text{fake}},p_{\text{real}})=\int|p_{\text{fake}}(\bm{x})-p_{\text{real}}(\bm{x})|d\bm{x}
\end{equation}
That is to say, when the discriminator is sufficiently rich and well trained, the theoretical optimum of Hinge loss upon convergence minimizes TVD.
When the support sets of fake and real distributions have minimal overlap, TVD provides two key advantages over reverse KL divergence:
\textbf{1) Symmetry}: TVD yields the same discrepancy measure regardless of the initial distribution, whereas the asymmetric reverse KL may exhibit mode-seeking behavior and neglect other portions of the overall distribution.
For example, as illustrated in \cref{fig:kl}(c), TVD maintains substantial loss values and provides optimization directions covering the mode when $p_{\text{fake}}(\bm{x})\rightarrow 0$ while $p_{\text{real}}(\bm{x})>0$, whereas the reverse KL divergence suffers from gradient vanishing issues in such scenarios, as discussed in \cref{sec:importance_pretrain}.
\textbf{2) Boundedness}: TVD is bounded within [0,1], and thus mitigating disruptions from outliers during training, especially in our high-dimensional multimodal text-conditioned image and video distributions, avoiding the numerical instability inherent to reverse KL divergence caused by gradient exploding.




\section{Experiments}
\label{sec:experiments}

\begin{table}[t]
    \centering
    \footnotesize
    \begin{tabular}{m{2.1cm}m{0.2cm}m{0.2cm}*{4}{m{0.7cm}}}
        \toprule[1.5pt]
        
        \multirow{1}[2]{*}{\makebox[2.1cm][l]{\textbf{Method}}} & 
        \multirow{1}[2]{*}{\makebox[0.2cm][c]{\textbf{Step}}} &
        \multirow{1}[2]{*}{\makebox[0.2cm][c]{\textbf{NFE}}} &
        \makebox[0.7cm][c]{\makecell[c]{\textbf{CLIP}\\\textbf{Score}}} & 
        \makebox[0.7cm][c]{\makecell[c]{\textbf{Pick}\\\textbf{Score}}} & 
        \multirow{1}[2]{*}{\makebox[0.7cm][c]{\textbf{HPSv2}}} & 
        \multirow{1}[2]{*}{\makebox[0.7cm][c]{\textbf{MPS}}} \\
        
        \midrule[1.5pt]


        \makebox[2.1cm][l]{ADD~\cite{sauer2023adversarial} (512px)} & 
        \makebox[0.2cm][c]{1} &
        \makebox[0.2cm][c]{1} &
        \makebox[0.7cm][c]{{35.0088}} & 
        \makebox[0.7cm][c]{{22.1524}} & 
        \makebox[0.7cm][c]{{27.0971}} & 
        \makebox[0.7cm][c]{{10.4340}} \\

        \makebox[2.1cm][l]{LCM~\cite{luo2023latent}} & 
        \makebox[0.2cm][c]{1} &
        \makebox[0.2cm][c]{2} &
        \makebox[0.7cm][c]{{28.4669}} & 
        \makebox[0.7cm][c]{{20.1267}} & 
        \makebox[0.7cm][c]{{23.8246}} & 
        \makebox[0.7cm][c]{{4.8134}} \\

        \makebox[2.1cm][l]{Lightning~\cite{lin2024sdxllightning}} & 
        \makebox[0.2cm][c]{1} &
        \makebox[0.2cm][c]{1} &
        \makebox[0.7cm][c]{{33.4985}} & 
        \makebox[0.7cm][c]{{21.9194}} & 
        \makebox[0.7cm][c]{{27.1557}} & 
        \makebox[0.7cm][c]{{10.2285}} \\

        \makebox[2.1cm][l]{DMD2~\cite{yin2024improved}} & 
        \makebox[0.2cm][c]{1} &
        \makebox[0.2cm][c]{1} &
        \makebox[0.7cm][c]{35.2153} & 
        \makebox[0.7cm][c]{22.0978} & 
        \makebox[0.7cm][c]{27.4523} & 
        \makebox[0.7cm][c]{10.6947} \\

        \makebox[2.1cm][l]{\textbf{DMDX (Ours)}} & 
        \makebox[0.2cm][c]{1} &
        \makebox[0.2cm][c]{1} &
        \makebox[0.7cm][c]{\textbf{35.2557}} & 
        \makebox[0.7cm][c]{\textbf{22.2736}} & 
        \makebox[0.7cm][c]{\textbf{27.7046}} & 
        \makebox[0.7cm][c]{\textbf{11.1978}} \\

        \makebox[2.1cm][l]{\textcolor{gray}{SDXL-Base~\cite{rombach2022highresolution}}} & 
        \makebox[0.2cm][c]{\textcolor{gray}{25}} &
        \makebox[0.2cm][c]{\textcolor{gray}{50}} &
        \makebox[0.7cm][c]{\textcolor{gray}{35.0309}} & 
        \makebox[0.7cm][c]{\textcolor{gray}{22.2494}} & 
        \makebox[0.7cm][c]{\textcolor{gray}{27.3743}} & 
        \makebox[0.7cm][c]{\textcolor{gray}{10.7042}} \\

        \bottomrule[1.5pt]
    \end{tabular}
    \vspace{-0.2cm}
    \caption{
        Quantitative results on fully fine-tuning SDXL-Base.
    }
    \vspace{-0.25cm}
    \label{tab:sdxl_1step}
\end{table}
\begin{table}[t]
    \centering
    \footnotesize
    \begin{tabular}{m{2.1cm}m{0.2cm}m{0.2cm}*{4}{m{0.7cm}}}
        \toprule[1.5pt]
        \multirow{1}[2]{*}{\makebox[2.1cm][l]{\textbf{Method}}} & 
        \multirow{1}[2]{*}{\makebox[0.2cm][c]{\textbf{Step}}} &
        \multirow{1}[2]{*}{\makebox[0.2cm][c]{\textbf{NFE}}} &
        \makebox[0.7cm][c]{\makecell[c]{\textbf{CLIP}\\\textbf{Score}}} & 
        \makebox[0.7cm][c]{\makecell[c]{\textbf{Pick}\\\textbf{Score}}} & 
        \multirow{1}[2]{*}{\makebox[0.7cm][c]{\textbf{HPSv2}}} & 
        \multirow{1}[2]{*}{\makebox[0.7cm][c]{\textbf{MPS}}} \\
        
        \midrule[1.5pt]


        \makebox[2.1cm][l]{TSCD~\cite{sauer2023adversarial}} & 
        \makebox[0.2cm][c]{4} &
        \makebox[0.2cm][c]{8} &
        \makebox[0.7cm][c]{{34.0185}} & 
        \makebox[0.7cm][c]{{21.9665}} & 
        \makebox[0.7cm][c]{{27.2728}} & 
        \makebox[0.7cm][c]{{10.8600}} \\

        \makebox[2.1cm][l]{PCM~\cite{wang2024phased} (Shift=1)} & 
        \makebox[0.2cm][c]{4} &
        \makebox[0.2cm][c]{4} &
        \makebox[0.7cm][c]{{33.5042}} & 
        \makebox[0.7cm][c]{{21.9703}} & 
        \makebox[0.7cm][c]{{27.3680}} & 
        \makebox[0.7cm][c]{{10.5707}} \\

        \makebox[2.1cm][l]{PCM~\cite{wang2024phased} (Shift=3)} & 
        \makebox[0.2cm][c]{4} &
        \makebox[0.2cm][c]{4} &
        \makebox[0.7cm][c]{{33.3818}} & 
        \makebox[0.7cm][c]{{21.9396}} & 
        \makebox[0.7cm][c]{{27.1146}} & 
        \makebox[0.7cm][c]{{10.5635}} \\

        \makebox[2.1cm][l]{PCM~\cite{wang2024phased} (Stoch.)} & 
        \makebox[0.2cm][c]{4} &
        \makebox[0.2cm][c]{4} &
        \makebox[0.7cm][c]{{33.4185}} & 
        \makebox[0.7cm][c]{{21.8822}} & 
        \makebox[0.7cm][c]{{27.3177}} & 
        \makebox[0.7cm][c]{{10.5200}} \\

        \makebox[2.1cm][l]{Flash~\cite{chadebec2024flash}} & 
        \makebox[0.2cm][c]{4} &
        \makebox[0.2cm][c]{4} &
        \makebox[0.7cm][c]{{34.3978}} & 
        \makebox[0.7cm][c]{{22.0904}} & 
        \makebox[0.7cm][c]{{27.2586}} & 
        \makebox[0.7cm][c]{{10.6634}} \\

        \makebox[2.1cm][l]{\textbf{ADM (Ours)}} & 
        \makebox[0.2cm][c]{4} &
        \makebox[0.2cm][c]{4} &
        \makebox[0.7cm][c]{\textbf{34.9076}} & 
        \makebox[0.7cm][c]{\textbf{22.5471}} & 
        \makebox[0.7cm][c]{\textbf{28.4492}} & 
        \makebox[0.7cm][c]{\textbf{11.9543}} \\

        \makebox[2.1cm][l]{\textcolor{gray}{SD3-Medium~\cite{esser2024scaling}}} & 
        \makebox[0.2cm][c]{\textcolor{gray}{25}} &
        \makebox[0.2cm][c]{\textcolor{gray}{50}} &
        \makebox[0.7cm][c]{\textcolor{gray}{34.7633}} & 
        \makebox[0.7cm][c]{\textcolor{gray}{22.2961}} & 
        \makebox[0.7cm][c]{\textcolor{gray}{27.9733}} & 
        \makebox[0.7cm][c]{\textcolor{gray}{11.3652}} \\

        \midrule[0.75pt]


        \makebox[2.1cm][l]{LADD~\cite{sauer2024fast}} & 
        \makebox[0.2cm][c]{4} &
        \makebox[0.2cm][c]{4} &
        \makebox[0.7cm][c]{{34.7395}} & 
        \makebox[0.7cm][c]{{22.3958}} & 
        \makebox[0.7cm][c]{{27.4923}} & 
        \makebox[0.7cm][c]{{11.4372}} \\

        \makebox[2.1cm][l]{\textbf{ADM (Ours)}} & 
        \makebox[0.2cm][c]{4} &
        \makebox[0.2cm][c]{4} &
        \makebox[0.7cm][c]{\textbf{34.9730}} & 
        \makebox[0.7cm][c]{\textbf{22.8842}} & 
        \makebox[0.7cm][c]{\textbf{27.7331}} & 
        \makebox[0.7cm][c]{\textbf{12.2350}} \\

        \makebox[2.1cm][l]{\textcolor{gray}{SD3.5-Large~\cite{esser2024scaling}}} & 
        \makebox[0.2cm][c]{\textcolor{gray}{25}} &
        \makebox[0.2cm][c]{\textcolor{gray}{50}} &
        \makebox[0.7cm][c]{\textcolor{gray}{34.9668}} & 
        \makebox[0.7cm][c]{\textcolor{gray}{22.5087}} & 
        \makebox[0.7cm][c]{\textcolor{gray}{27.9688}} & 
        \makebox[0.7cm][c]{\textcolor{gray}{11.5826}} \\

        \bottomrule[1.5pt]
    \end{tabular}
    \vspace{-0.2cm}
    \caption{
        Quantitative results on LoRA fine-tuning SD3-Medium and fully fine-tuning SD3.5-Large.
    }
    \vspace{-0.25cm}
    \label{tab:sd3_4step}
\end{table}
\begin{table}[t]
    \centering
    \footnotesize
    \begin{tabular}{m{2.8cm}m{0.3cm}m{0.3cm}*{3}{m{0.8cm}}}
        \toprule[1.5pt]
        \multirow{1}[2]{*}{\makebox[2.8cm][l]{\textbf{Method}}} & 
        \multirow{1}[2]{*}{\makebox[0.2cm][c]{\textbf{Step}}} &
        \multirow{1}[2]{*}{\makebox[0.2cm][c]{\textbf{NFE}}} &
        \makebox[0.8cm][c]{\makecell[c]{\textbf{Final}\\\textbf{Score}}} & 
        \makebox[0.8cm][c]{\makecell[c]{\textbf{Quality}\\\textbf{Score}}} & 
        \makebox[0.8cm][c]{\makecell[c]{\textbf{Semantic}\\\textbf{Score}}} \\
        
        \midrule[1.5pt]

        \makebox[2.8cm][l]{\textbf{ADM}} & 
        \makebox[0.2cm][c]{8} &
        \makebox[0.2cm][c]{8} &
        \makebox[0.8cm][c]{78.584} & 
        \makebox[0.8cm][c]{80.825} & 
        \makebox[0.8cm][c]{69.621} \\

        \makebox[2.8cm][l]{+ Longer Training $\times$2} & 
        \makebox[0.2cm][c]{8} &
        \makebox[0.2cm][c]{8} &
        \makebox[0.8cm][c]{80.764} & 
        \makebox[0.8cm][c]{\textbf{83.031}} & 
        \makebox[0.8cm][c]{71.693} \\

        \makebox[2.8cm][l]{\textbf{ADM w/ CFG}} & 
        \makebox[0.2cm][c]{8} &
        \makebox[0.2cm][c]{16} &
        \makebox[0.8cm][c]{79.865} & 
        \makebox[0.8cm][c]{80.938} & 
        \makebox[0.8cm][c]{75.569} \\

        \makebox[2.8cm][l]{+ Longer Training $\times$2} & 
        \makebox[0.2cm][c]{8} &
        \makebox[0.2cm][c]{16} &
        \makebox[0.8cm][c]{\textbf{81.796}} & 
        \makebox[0.8cm][c]{83.008} & 
        \makebox[0.8cm][c]{\textbf{76.947}} \\

        \makebox[2.8cm][l]{\textcolor{gray}{CogVideoX-2b~\cite{yang2024cogvideox}}} & 
        \makebox[0.2cm][c]{\textcolor{gray}{100}} &
        \makebox[0.2cm][c]{\textcolor{gray}{200}} &
        \makebox[0.8cm][c]{\textcolor{gray}{80.036}} & 
        \makebox[0.8cm][c]{\textcolor{gray}{80.801}} & 
        \makebox[0.8cm][c]{\textcolor{gray}{76.974}} \\

        \midrule[0.75pt]

        \makebox[2.8cm][l]{\textbf{ADM}} & 
        \makebox[0.2cm][c]{8} &
        \makebox[0.2cm][c]{8} &
        \makebox[0.8cm][c]{\textbf{82.067}} & 
        \makebox[0.8cm][c]{\textbf{83.227}} & 
        \makebox[0.8cm][c]{\textbf{77.423}} \\

        \makebox[2.8cm][l]{\textbf{ADM w/ CFG}} & 
        \makebox[0.2cm][c]{8} &
        \makebox[0.2cm][c]{16} &
        \makebox[0.8cm][c]{80.982} & 
        \makebox[0.8cm][c]{82.165} & 
        \makebox[0.8cm][c]{76.251} \\

        \makebox[2.8cm][l]{\textcolor{gray}{CogVideoX-5b~\cite{yang2024cogvideox}}} & 
        \makebox[0.2cm][c]{\textcolor{gray}{100}} &
        \makebox[0.2cm][c]{\textcolor{gray}{200}} &
        \makebox[0.8cm][c]{\textcolor{gray}{81.226}} & 
        \makebox[0.8cm][c]{\textcolor{gray}{81.785}} & 
        \makebox[0.8cm][c]{\textcolor{gray}{78.987}} \\

        \bottomrule[1.5pt]
    \end{tabular}
    \vspace{-0.2cm}
    \caption{
        Quantitative results on fully fine-tuning CogVideoX.
    }
    \vspace{-0.45cm}
    \label{tab:vbench_short}
\end{table}

\noindent\textbf{Models.}
For one-step distillation, we employ both Adversarial Distillation Pre-training (ADP) and ADM fine-tuning on SDXL-Base~\cite{rombach2022highresolution}, termed \ul{{DMDX}}.
For multi-step distillation, we only employ ADM training on both text-to-image models SD3-Medium, SD3.5-Large~\cite{esser2024scaling}, and text-to-video models CogVideoX-2b, CogVideoX-5b~\cite{yang2024cogvideox}.
Following most concurrent works, we didn't use Classifier-Free Guidance (CFG)~\cite{ho2021classifierfree} in our text-to-image model.
We tried conducting CFG-integrated experiments on the text-to-video model, with the approach detailed in \cref{exp:fewstep_video}.

\noindent\textbf{Datasets.}
No visual data is required for both ADP and ADM proposed in this work.
For image generators, we utilize the text prompts from JourneyDB~\cite{sun2023journeydb} that exhibits high-level of detail and specificity for training.
For video generators, we collect training prompts from OpenVid-1M~\cite{nan2024openvid1m}, Vript~\cite{yang2024vript} and Open-Sora-Plan-v1.1.0~\cite{pku-yuanlab2024opensoraplanv110}.

\noindent\textbf{Evaluation.}
The image generators are evaluated on 10K prompts from COCO 2014~\cite{lin2014microsoft} following DMD2~\cite{yin2024improved}. 
We report the CLIP score~\cite{radford2021learning} and human preference benchmarks PickScore~\cite{kirstain2023pickapic}, HPSv2~\cite{wu2023human} and MPS~\cite{zhang2024learning} as many concurrent works including Hyper-SD~\cite{ren2024hypersd} and Emu3~\cite{wang2024emu3}.
But we don't include Hyper-SD in the one-step quantitative comparison because one-step Hyper-SDXL has been optimized directly on human feedback with ReFL~\cite{xu2023imagereward}.
Instead, we compare with the TSCD algorithm proposed therein on SD3-Medium~\cite{hu2022lora}, since 4-step Hyper-SD3 LoRA is without ReFL optimization.
The video generators are evaluated on VBench~\cite{huang2024vbench}, which consists of multiple quality and semantic dimensions comprehensively.

\noindent\textbf{Hyperparameters.}
Despite having multiple models to train in our proposed ADP and ADM, we achieve satisfactory visual fidelity and structural integrity without extensive hyperparameter tuning.
For the remainder of this work, we only adjust the learning rate of generator for different experiments.
The optimizer settings for discriminators and fake models are uniform across all the experiments.
Unless longer training is specifically noted, we train the same for only 8K iterations with a batch size of 128 and 8 for text-to-image and text-to-video models, respectively. 
More specific implementation details are provided in \cref{appendix:implementation_details}.


\subsection{Efficient Image Synthesis}\label{exp:reflow_image}

\begin{figure}
    \centering
    \includegraphics[width=\linewidth]{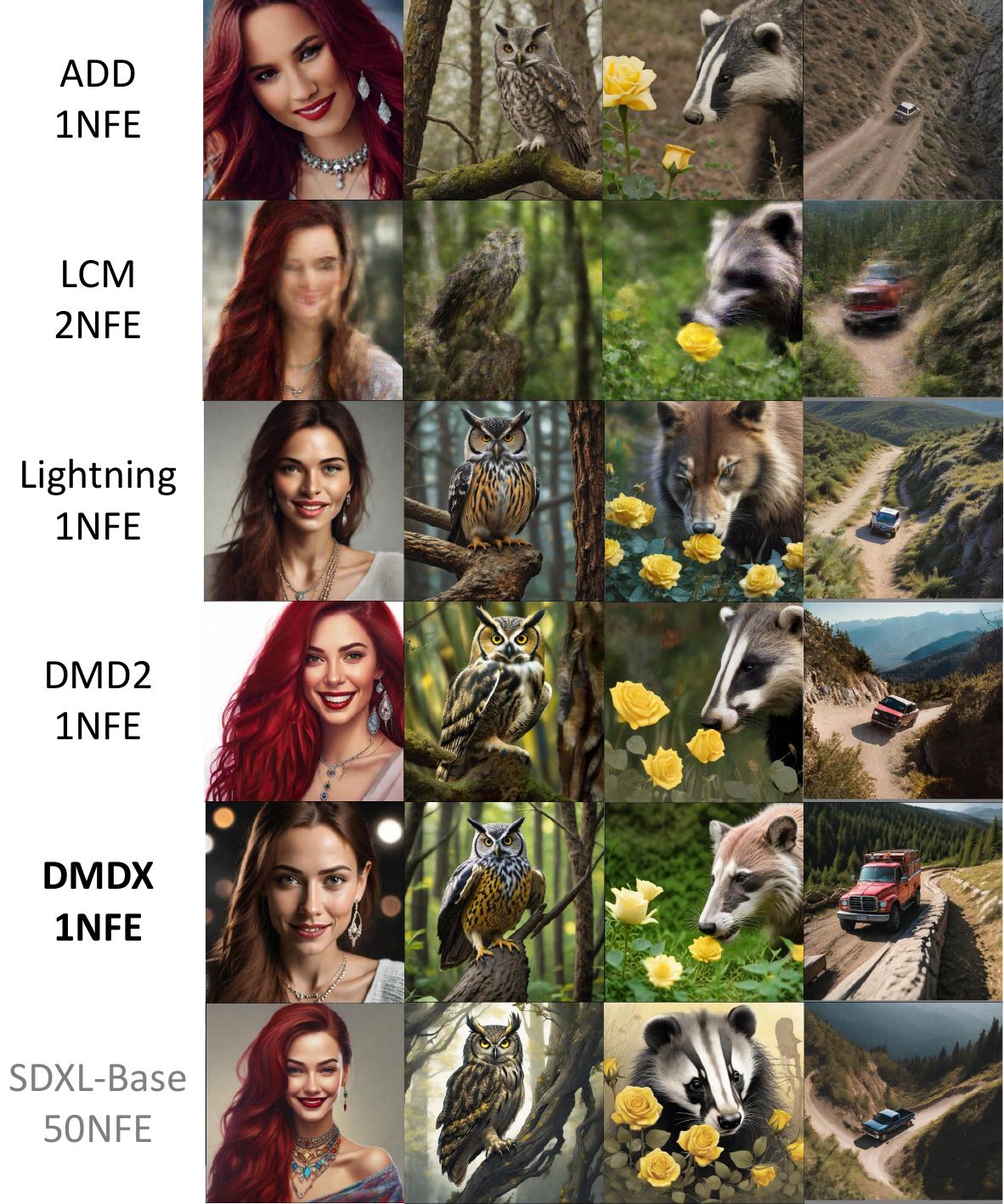}
    \vspace{-0.6cm}
    \caption{
    Qualitative results on fully fine-tuning SDXL-Base.
    }
    \vspace{-0.3cm}
    \label{fig:qualitative_sdxl}
\end{figure}

\cref{tab:sdxl_1step} quantitatively compares our two-stage approach combing ADP and single-step ADM distillation with existing one-step distillation methods on fully fine-tuning SDXL-Base.
The results show that our method achieves excellent performance on both image-text alignment and human preference, which is consistent with the qualitative comparisons in \cref{fig:qualitative_sdxl}, including better portrait aesthetic, animal hair details, subject-background separation and physical structure.

For multi-step ADM distillation, it can serve as a standalone score distillation method.
We tried both fully fine-tuning and LoRA fine-tuning~\cite{hu2022lora} configurations and the quantitative results in \cref{tab:sd3_4step} demonstrate our superior performance.
Qualitative results are provided in \cref{appendix:main_results}.

\begin{table}[t]
    \centering
    \footnotesize
    \begin{tabular}{m{3.0cm}*{4}{m{0.7cm}}}
        \toprule[1.5pt]
        
        \multirow{1}[2]{*}{\makebox[3.0cm][l]{\textbf{Ablation}}} & 
        \makebox[0.7cm][c]{\makecell[c]{\textbf{CLIP}\\\textbf{Score}}} & 
        \makebox[0.7cm][c]{\makecell[c]{\textbf{Pick}\\\textbf{Score}}} & 
        \multirow{1}[2]{*}{\makebox[0.7cm][c]{\textbf{HPSv2}}} & 
        \multirow{1}[2]{*}{\makebox[0.7cm][c]{\textbf{MPS}}} \\
        
        \midrule[1.5pt]

        \rowcolor[gray]{0.9} \multicolumn{5}{l}{\textit{Ablation on adversarial distillation.}} \\

        \makebox[3.0cm][l]{{\ttfamily\detokenize{A1}}: Rectified Flow~\cite{liu2022flow}} & 
        \makebox[0.7cm][c]{27.4376} & 
        \makebox[0.7cm][c]{20.0211} & 
        \makebox[0.7cm][c]{23.6093} & 
        \makebox[0.7cm][c]{4.4518} \\

        \makebox[3.0cm][l]{{\ttfamily\detokenize{A2}}: DINOv2 as pixel-space} & 
        \makebox[0.7cm][c]{34.1836} & 
        \makebox[0.7cm][c]{21.8750} & 
        \makebox[0.7cm][c]{27.1039} & 
        \makebox[0.7cm][c]{10.2407} \\

        \makebox[3.0cm][l]{{\ttfamily\detokenize{A3}}: $\lambda_1=0.7,\lambda_2=0.3$} & 
        \makebox[0.7cm][c]{33.6943} & 
        \makebox[0.7cm][c]{21.6344} & 
        \makebox[0.7cm][c]{26.8902} & 
        \makebox[0.7cm][c]{9.9633} \\

        \makebox[3.0cm][l]{{\ttfamily\detokenize{A4}}: $\lambda_1=1.0,\lambda_2=0.0$} & 
        \makebox[0.7cm][c]{33.8929} & 
        \makebox[0.7cm][c]{21.7395} & 
        \makebox[0.7cm][c]{26.7869} & 
        \makebox[0.7cm][c]{10.0757} \\

        \makebox[3.0cm][l]{{\ttfamily\detokenize{A5}}: \textbf{w/o ADM (ADP only)}} & 
        \makebox[0.7cm][c]{\textbf{35.7723}} & 
        \makebox[0.7cm][c]{22.0095} & 
        \makebox[0.7cm][c]{27.3499} & 
        \makebox[0.7cm][c]{10.6646} \\

        \midrule[0.75pt]

        \rowcolor[gray]{0.9} \multicolumn{5}{l}{\textit{Ablation on score distillation.}} \\

        \makebox[3.0cm][l]{{\ttfamily\detokenize{B1}}: ADM w/o ADP} & 
        \makebox[0.7cm][c]{32.5020} & 
        \makebox[0.7cm][c]{21.7631} & 
        \makebox[0.7cm][c]{26.8732} & 
        \makebox[0.7cm][c]{10.8986} \\

        \makebox[3.0cm][l]{{\ttfamily\detokenize{B2}}: DMD Loss w/o ADP} & 
        \makebox[0.7cm][c]{32.7482} & 
        \makebox[0.7cm][c]{21.0341} & 
        \makebox[0.7cm][c]{25.9680} & 
        \makebox[0.7cm][c]{8.8977} \\


        \makebox[3.0cm][l]{{\ttfamily\detokenize{B3}}: DMD Loss w/ ADP} & 
        \makebox[0.7cm][c]{34.5119} & 
        \makebox[0.7cm][c]{21.9366} & 
        \makebox[0.7cm][c]{27.3985} & 
        \makebox[0.7cm][c]{10.6046} \\

        \makebox[3.0cm][l]{{\ttfamily\detokenize{B4}}: \textbf{DMDX (Ours)}} & 
        \makebox[0.7cm][c]{{35.2557}} & 
        \makebox[0.7cm][c]{\textbf{22.2736}} & 
        \makebox[0.7cm][c]{\textbf{27.7046}} & 
        \makebox[0.7cm][c]{\textbf{11.1978}} \\

        \bottomrule[1.5pt]
    \end{tabular}
    \vspace{-0.2cm}
    \caption{
        Quantitative results on ablation studies.
    }
    \vspace{-0.45cm}
    \label{tab:ablation_dmdx}
\end{table}

\subsection{Efficient Video Synthesis}\label{exp:fewstep_video}


As quantitatively shown in \cref{tab:vbench_short}, except for the regular 8-step ADM distillation for both sizes of CogVideoX, we also try integrating Classifier-Free Guidance (CFG)~\cite{ho2021classifierfree} for text-to-video task.
Specifically, we assign the CFG scale for real model by randomly sampling within the range $[5.0,7.0]$, while assigning the few-step generator's scale through explicit subtraction of 2.0 from the real model's value.
Empirically, we found that this subtractive offset requires progressive enlargement as the targeted sampling steps decreases.
There is no CFG for the fake model.
The VBench~\cite{huang2024vbench} results demonstrate that our few-step generators achieve comparable performance to the base model with 92-96\% acceleration.
We conduct additional evaluations on the longer training of 2B model, motivated by the observation that DMD loss did't converge well at 8K iterations as \cref{fig:ablation_relation_dmd_dmdx} indicates.
Experimental results demonstrate that the DMD loss is also approximately optimized by the learnable discriminator during ADM distillation.
More quantitative and qualitative results refer to \cref{appendix:main_results}.


\subsection{Ablation Studies}\label{sec:ablation}

In \cref{tab:ablation_dmdx}, we conduct extensive ablation studies on fully fine-tuning SDXL-Base to validate our effectiveness.
Qualitative comparisons are provided in \cref{appendix:main_results}.


\noindent\textbf{Effect of ADP.}
The following conclusions can be drawn: \textbf{1)} Single reflow process provides very limited effectiveness ({\ttfamily\detokenize{A1}}), while multiple processes achieving effective rectification requires considerable computational expense.
\textbf{2)} Using SAM~\cite{kirillov2023segment} offers more imaging fidelity compared to the widely adopted DINOv2~\cite{oquab2024dinov2} ({\ttfamily\detokenize{A2/A5}}), which is likely due to SAM's higher resolution of 1024px (compared to DINOv2's 518px).
\textbf{3)} The weighting $\lambda_2$ for pixel space should neither be too large nor too small. Excessive weighting leads to degraded structural integrity ({\ttfamily\detokenize{A3/A5}}), while insufficient weighting results in blurriness ({\ttfamily\detokenize{A4/A5}}).
Qualitative results in \cref{appendix:main_results} provide more visible distinction.

\noindent\textbf{Effect of ADM.}
We summarize the key findings:
\textbf{1)} The absence of ADP exerts substantial performance degradation ({\ttfamily\detokenize{B1/B4}}), which is aligned with our analysis in \cref{sec:importance_pretrain}.
\textbf{2)} Without a regularizer, the DMD loss underperforms compared to standalone ADM ({\ttfamily\detokenize{B1/B2}}), indicating its poor robustness.
\textbf{3)} Although DMD loss optimization also benefits from ADP ({\ttfamily\detokenize{B2/B3}}), its distribution matching capability remains inferior to ADM ({\ttfamily\detokenize{B3/B4}}).

\begin{table}[t]
    \centering
    \footnotesize
    \begin{tabular}{m{1cm}m{1cm}*{4}{m{0.7cm}}}
        \toprule[1.5pt]
        
        \multirow{1}[2]{*}{\makebox[1cm][c]{\textbf{TTUR}}} & 
        \makebox[1cm][c]{\makecell[c]{\textbf{Training}\\\textbf{Time}}} & 
        \makebox[0.7cm][c]{\makecell[c]{\textbf{CLIP}\\\textbf{Score}}} & 
        \makebox[0.7cm][c]{\makecell[c]{\textbf{Pick}\\\textbf{Score}}} & 
        \multirow{1}[2]{*}{\makebox[0.7cm][c]{\textbf{HPSv2}}} & 
        \multirow{1}[2]{*}{\makebox[0.7cm][c]{\textbf{MPS}}} \\
        
        \midrule[1.5pt]

        \makebox[1cm][c]{{1}} & 
        \makebox[1cm][c]{{$\times$1.00}} & 
        \makebox[0.7cm][c]{35.2557} & 
        \makebox[0.7cm][c]{{22.2736}} & 
        \makebox[0.7cm][c]{27.7046} & 
        \makebox[0.7cm][c]{{11.1978}} \\

        \makebox[1cm][c]{4} & 
        \makebox[1cm][c]{$\times$1.85} & 
        \makebox[0.7cm][c]{35.2583} & 
        \makebox[0.7cm][c]{22.2773} & 
        \makebox[0.7cm][c]{27.7255} & 
        \makebox[0.7cm][c]{11.2720} \\

        \makebox[1cm][c]{8} & 
        \makebox[1cm][c]{$\times$2.53} & 
        \makebox[0.7cm][c]{35.3299} & 
        \makebox[0.7cm][c]{22.2883} & 
        \makebox[0.7cm][c]{27.7586} & 
        \makebox[0.7cm][c]{11.2838} \\

        \bottomrule[1.5pt]
    \end{tabular}
    \vspace{-0.2cm}
    \caption{
        Ablation study on two time-scale update rule.
    }
    \vspace{-0.2cm}
    \label{tab:ablation_ttur}
\end{table}
\begin{table}[t]
    \centering
    \footnotesize
    \begin{tabular}{c*{6}{m{0.7cm}}}
        \toprule[1.5pt]

         & \makebox[0.7cm][c]{ADD} & \makebox[0.7cm][c]{LCM} & \makebox[0.7cm][c]{Lightning} & \makebox[0.7cm][c]{DMD2} & \makebox[0.7cm][c]{\textbf{Ours}} & \makebox[0.7cm][c]{\textcolor{gray}{Teacher}} \\
        
        \midrule[1.5pt]

        \textbf{LPIPS$\uparrow$} & \makebox[0.7cm][c]{0.6071} & \makebox[0.7cm][c]{0.6257} & \makebox[0.7cm][c]{0.6707} & \makebox[0.7cm][c]{0.6715} & \makebox[0.7cm][c]{\textbf{0.7156}} & \makebox[0.7cm][c]{\textcolor{gray}{0.6936}} \\

        \bottomrule[1.5pt]
    \end{tabular}
    \vspace{-0.2cm}
    \caption{
        Quantitative diversity evaluation on PartiPrompts~\cite{yu2022scaling}.
    }
    \vspace{-0.45cm}
    \label{tab:ablation_diversity}
\end{table}

\noindent\textbf{Effect of TTUR.}
\cref{tab:ablation_ttur} presents the impact of different TTUR settings on final performance and training duration. 
The results demonstrate that increasing TTUR yields only marginal performance gains while nearly doubling training time, rendering this trade-off clearly unwarranted.
This highlights the critical role of our proposed ADP in one-step distillation and suggests that the training instability in DMD2 likely stems from insufficient support set overlap.

\noindent\textbf{Diversity evaluation.}
Following DMD2~\cite{yin2024improved}, we generate 4 samples per prompt on Partiprompts~\cite{yu2022scaling} with different seeds, and report the averaged pairwise LPIPS similarities~\cite{zhang2018unreasonable} in \cref{tab:ablation_diversity}.
The results indicate that our method significantly outperforms others in diversity.
More randomly curated multi-seed samples can be found in \cref{appendix:main_results}.

\section{Limitations}
\label{sec:limirations}


We realize that one weakness is that teacher model might require CFG to produce accurate score prediction. 
Our experiments suggest this is a common characteristic of score distillation methods generally, not a limitation unique to our approach. 
This limits the application to guidance-distilled models such as FLUX.1-dev~\cite{blackforestlabs2024flux1dev}, which could be a promising research topic for future work.

\section*{Acknowledgments}
This work was supported in part by the National Natural Science Foundation of China (U22A2095, 12326618, 62276281), Guangdong Basic and Applied Basic Research Foundation, China (2024A1515011882) and the Project of Guangdong Provincial Key Laboratory of Information Security Technology (2023B1212060026).

{
    \small
    \bibliographystyle{ieeenat_fullname}
    \bibliography{main}

\begin{thebibliography}{94}
\providecommand{\natexlab}[1]{#1}
\providecommand{\url}[1]{\texttt{#1}}
\expandafter\ifx\csname urlstyle\endcsname\relax
  \providecommand{\doi}[1]{doi: #1}\else
  \providecommand{\doi}{doi: \begingroup \urlstyle{rm}\Url}\fi

\bibitem[{Black Forest Labs}(2024)]{blackforestlabs2024flux1dev}
{Black Forest Labs}.
\newblock Flux.1-dev.
\newblock \url{https://huggingface.co/black-forest-labs/FLUX.1-dev}, 2024.

\bibitem[Bynagari(2017)]{bynagari2017gans}
Naresh~Babu Bynagari.
\newblock Gans trained by a two time-scale update rule converge to a local nash equilibrium.
\newblock In \emph{NeurIPS}, 2017.

\bibitem[Chadebec et~al.(2024)Chadebec, Tasar, Benaroche, and Aubin]{chadebec2024flash}
Clement Chadebec, Onur Tasar, Eyal Benaroche, and Benjamin Aubin.
\newblock Flash diffusion: Accelerating any conditional diffusion model for few steps image generation.
\newblock \emph{arXiv preprint arXiv:2406.02347}, 2024.

\bibitem[Chen et~al.(2016)Chen, Xu, Zhang, and Guestrin]{chen2016training}
Tianqi Chen, Bing Xu, Chiyuan Zhang, and Carlos Guestrin.
\newblock Training deep nets with sublinear memory cost.
\newblock \emph{arXiv preprint arXiv:1604.06174}, 2016.

\bibitem[Dosovitskiy et~al.(2021)Dosovitskiy, Beyer, Kolesnikov, Weissenborn, Zhai, Unterthiner, Dehghani, Minderer, Heigold, Gelly, Uszkoreit, and Houlsby]{dosovitskiy2021image}
Alexey Dosovitskiy, Lucas Beyer, Alexander Kolesnikov, Dirk Weissenborn, Xiaohua Zhai, Thomas Unterthiner, Mostafa Dehghani, Matthias Minderer, Georg Heigold, Sylvain Gelly, Jakob Uszkoreit, and Neil Houlsby.
\newblock An image is worth 16x16 words: Transformers for image recognition at scale.
\newblock In \emph{ICLR}, 2021.

\bibitem[Esser et~al.(2024)Esser, Kulal, Blattmann, Entezari, M{\"u}ller, Saini, Levi, Lorenz, Sauer, Boesel, Podell, Dockhorn, English, Lacey, Goodwin, Marek, and Rombach]{esser2024scaling}
Patrick Esser, Sumith Kulal, Andreas Blattmann, Rahim Entezari, Jonas M{\"u}ller, Harry Saini, Yam Levi, Dominik Lorenz, Axel Sauer, Frederic Boesel, Dustin Podell, Tim Dockhorn, Zion English, Kyle Lacey, Alex Goodwin, Yannik Marek, and Robin Rombach.
\newblock Scaling rectified flow transformers for high-resolution image synthesis.
\newblock In \emph{ICML}, 2024.

\bibitem[Geng et~al.(2025)Geng, Pokle, Luo, Lin, and Kolter]{geng2024consistency}
Zhengyang Geng, Ashwini Pokle, William Luo, Justin Lin, and J~Zico Kolter.
\newblock Consistency models made easy.
\newblock In \emph{ICLR}, 2025.

\bibitem[Goodfellow et~al.(2014)Goodfellow, {Pouget-Abadie}, Mirza, Xu, {Warde-Farley}, Ozair, Courville, and Bengio]{goodfellow2014generative}
Ian~J. Goodfellow, Jean {Pouget-Abadie}, Mehdi Mirza, Bing Xu, David {Warde-Farley}, Sherjil Ozair, Aaron Courville, and Yoshua Bengio.
\newblock Generative adversarial nets.
\newblock In \emph{NeurIPS}, 2014.

\bibitem[He et~al.(2016)He, Zhang, Ren, and Sun]{he2016deep}
Kaiming He, Xiangyu Zhang, Shaoqing Ren, and Jian Sun.
\newblock Deep residual learning for image recognition.
\newblock In \emph{CVPR}, pages 770--778, 2016.

\bibitem[Hendrycks and Gimpel(2016)]{hendrycks2016gaussian}
Dan Hendrycks and Kevin Gimpel.
\newblock Gaussian error linear units (gelus).
\newblock \emph{arXiv preprint arXiv:1606.08415}, 2016.

\bibitem[Ho and Salimans(2021)]{ho2021classifierfree}
Jonathan Ho and Tim Salimans.
\newblock Classifier-free diffusion guidance.
\newblock In \emph{NeurIPS Workshops}, 2021.

\bibitem[Ho et~al.(2020)Ho, Jain, and Abbeel]{ho2020denoising}
Jonathan Ho, Ajay Jain, and Pieter Abbeel.
\newblock Denoising diffusion probabilistic models.
\newblock In \emph{NeurIPS}, 2020.

\bibitem[Hu et~al.(2022)Hu, Shen, Wallis, {Allen-Zhu}, Li, Wang, Wang, and Chen]{hu2022lora}
Edward~J. Hu, Yelong Shen, Phillip Wallis, Zeyuan {Allen-Zhu}, Yuanzhi Li, Shean Wang, Lu Wang, and Weizhu Chen.
\newblock Lora: Low-rank adaptation of large language models.
\newblock In \emph{ICLR}, 2022.

\bibitem[Huang et~al.(2024)Huang, He, Yu, Zhang, Si, Jiang, Zhang, Wu, Jin, Chanpaisit, Wang, Chen, Wang, Lin, Qiao, and Liu]{huang2024vbench}
Ziqi Huang, Yinan He, Jiashuo Yu, Fan Zhang, Chenyang Si, Yuming Jiang, Yuanhan Zhang, Tianxing Wu, Qingyang Jin, Nattapol Chanpaisit, Yaohui Wang, Xinyuan Chen, Limin Wang, Dahua Lin, Yu Qiao, and Ziwei Liu.
\newblock Vbench: Comprehensive benchmark suite for video generative models.
\newblock In \emph{CVPR}, 2024.

\bibitem[Jacobs et~al.(2023)Jacobs, Tanaka, Zhang, Zhang, Song, Rajbhandari, and He]{jacobs2023deepspeed}
Sam~Ade Jacobs, Masahiro Tanaka, Chengming Zhang, Minjia Zhang, Shuaiwen~Leon Song, Samyam Rajbhandari, and Yuxiong He.
\newblock Deepspeed ulysses: System optimizations for enabling training of extreme long sequence transformer models.
\newblock \emph{arXiv preprint arXiv:2309.14509}, 2023.

\bibitem[Jayashankar et~al.(2025)Jayashankar, Ryu, and Wornell]{jayashankar2025scoreofmixture}
Tejas Jayashankar, J.~Jon Ryu, and Gregory Wornell.
\newblock Score-of-mixture training: Training one-step generative models made simple via score estimation of mixture distributions.
\newblock \emph{arXiv preprint arXiv:2502.09609}, 2025.

\bibitem[Karras et~al.(2022)Karras, Aittala, Aila, and Laine]{karras2022elucidating}
Tero Karras, Miika Aittala, Timo Aila, and Samuli Laine.
\newblock Elucidating the design space of diffusion-based generative models.
\newblock In \emph{NeurIPS}, 2022.

\bibitem[Kim et~al.(2024)Kim, Lai, Liao, Murata, Takida, Uesaka, He, Mitsufuji, and Ermon]{kim2024consistency}
Dongjun Kim, Chieh-Hsin Lai, Wei-Hsiang Liao, Naoki Murata, Yuhta Takida, Toshimitsu Uesaka, Yutong He, Yuki Mitsufuji, and Stefano Ermon.
\newblock Consistency trajectory models: Learning probability flow ode trajectory of diffusion.
\newblock In \emph{ICLR}, 2024.

\bibitem[Kirillov et~al.(2023)Kirillov, Mintun, Ravi, Mao, Rolland, Gustafson, Xiao, Whitehead, Berg, Lo, Doll{\'a}r, and Girshick]{kirillov2023segment}
Alexander Kirillov, Eric Mintun, Nikhila Ravi, Hanzi Mao, Chloe Rolland, Laura Gustafson, Tete Xiao, Spencer Whitehead, Alexander~C. Berg, Wan-Yen Lo, Piotr Doll{\'a}r, and Ross Girshick.
\newblock Segment anything.
\newblock In \emph{ICCV}, pages 3992--4003, 2023.

\bibitem[Kirstain et~al.(2023)Kirstain, Polyak, Singer, Matiana, Penna, and Levy]{kirstain2023pickapic}
Yuval Kirstain, Adam Polyak, Uriel Singer, Shahbuland Matiana, Joe Penna, and Omer Levy.
\newblock Pick-a-pic: An open dataset of user preferences for text-to-image generation.
\newblock In \emph{NeuriPS}, 2023.

\bibitem[Kohler et~al.(2024)Kohler, Pumarola, Sch{\"o}nfeld, Sanakoyeu, Sumbaly, Vajda, and Thabet]{kohler2024imagine}
Jonas Kohler, Albert Pumarola, Edgar Sch{\"o}nfeld, Artsiom Sanakoyeu, Roshan Sumbaly, Peter Vajda, and Ali Thabet.
\newblock Imagine flash: Accelerating emu diffusion models with backward distillation.
\newblock \emph{arXiv preprint arXiv:2405.05224}, 2024.

\bibitem[Lim and Ye(2017)]{lim2017geometric}
Jae~Hyun Lim and Jong~Chul Ye.
\newblock Geometric gan.
\newblock \emph{arXiv preprint arXiv:1705.02894}, 2017.

\bibitem[Lin et~al.(2024)Lin, Wang, and Yang]{lin2024sdxllightning}
Shanchuan Lin, Anran Wang, and Xiao Yang.
\newblock Sdxl-lightning: Progressive adversarial diffusion distillation.
\newblock \emph{arXiv preprint arXiv:2402.13929}, 2024.

\bibitem[Lin et~al.(2025)Lin, Xia, Ren, Yang, Xiao, Jiang, and Seed]{lin2025diffusion}
Shanchuan Lin, Xin Xia, Yuxi Ren, Ceyuan Yang, Xuefeng Xiao, Lu Jiang, and ByteDance Seed.
\newblock Diffusion adversarial post-training for one-step video generation.
\newblock \emph{arXiv preprint arXiv:2501.08316}, 2025.

\bibitem[Lin et~al.(2014)Lin, Maire, Belongie, Bourdev, Girshick, Hays, Perona, Ramanan, Zitnick, and Doll{\'a}r]{lin2014microsoft}
Tsung-Yi Lin, Michael Maire, Serge Belongie, Lubomir Bourdev, Ross Girshick, James Hays, Pietro Perona, Deva Ramanan, C.~Lawrence Zitnick, and Piotr Doll{\'a}r.
\newblock Microsoft coco: Common objects in context.
\newblock In \emph{ECCV}, 2014.

\bibitem[Lipman et~al.(2023)Lipman, Chen, {Ben-Hamu}, Nickel, and Le]{lipman2023flow}
Yaron Lipman, Ricky T.~Q. Chen, Heli {Ben-Hamu}, Maximilian Nickel, and Matt Le.
\newblock Flow matching for generative modeling.
\newblock \emph{arXiv preprint arXiv:2210.02747}, 2023.

\bibitem[Liu et~al.(2022)Liu, Gong, and Liu]{liu2022flow}
Xingchao Liu, Chengyue Gong, and Qiang Liu.
\newblock Flow straight and fast: Learning to generate and transfer data with rectified flow.
\newblock In \emph{ICLR}, 2022.

\bibitem[Liu et~al.(2024)Liu, Zhang, Ma, Peng, and Liu]{liu2024instaflow}
Xingchao Liu, Xiwen Zhang, Jianzhu Ma, Jian Peng, and Qiang Liu.
\newblock Instaflow: One step is enough for high-quality diffusion-based text-to-image generation.
\newblock In \emph{ICLR}, 2024.

\bibitem[Loshchilov and Hutter(2019)]{loshchilov2019decoupled}
Ilya Loshchilov and Frank Hutter.
\newblock Decoupled weight decay regularization.
\newblock In \emph{ICLR}, 2019.

\bibitem[Lu and Song(2024)]{lu2024simplifying}
Cheng Lu and Yang Song.
\newblock Simplifying, stabilizing and scaling continuous-time consistency models.
\newblock \emph{arXiv preprint arXiv:2410.11081}, 2024.

\bibitem[Lu et~al.(2022)Lu, Zhang, Lin, Ma, Xie, and Lai]{lu2022improving}
Yanzuo Lu, Manlin Zhang, Yiqi Lin, Andy~J. Ma, Xiaohua Xie, and Jianhuang Lai.
\newblock Improving pre-trained masked autoencoder via locality enhancement for person re-identification.
\newblock In \emph{PRCV}, pages 509--521, 2022.

\bibitem[Lu et~al.(2024{\natexlab{a}})Lu, Shen, Ma, Xie, and Lai]{lu2024mlnet}
Yanzuo Lu, Meng Shen, Andy~J Ma, Xiaohua Xie, and Jian-Huang Lai.
\newblock Mlnet: Mutual learning network with neighborhood invariance for universal domain adaptation.
\newblock In \emph{AAAI}, pages 3900--3908, 2024{\natexlab{a}}.

\bibitem[Lu et~al.(2024{\natexlab{b}})Lu, Zhang, Ma, Xie, and Lai]{lu2024coarsefine}
Yanzuo Lu, Manlin Zhang, Andy~J Ma, Xiaohua Xie, and Jianhuang Lai.
\newblock Coarse-to-fine latent diffusion for pose-guided person image synthesis.
\newblock In \emph{CVPR}, pages 6420--6429, 2024{\natexlab{b}}.

\bibitem[Luo et~al.(2023{\natexlab{a}})Luo, Tan, Huang, Li, and Zhao]{luo2023latent}
Simian Luo, Yiqin Tan, Longbo Huang, Jian Li, and Hang Zhao.
\newblock Latent consistency models: Synthesizing high-resolution images with few-step inference.
\newblock \emph{arXiv preprint arXiv:2310.04378}, 2023{\natexlab{a}}.

\bibitem[Luo et~al.(2023{\natexlab{b}})Luo, Tan, Patil, Gu, von Platen, Passos, Huang, Li, and Zhao]{luo2023lcmlora}
Simian Luo, Yiqin Tan, Suraj Patil, Daniel Gu, Patrick von Platen, Apolin{\'a}rio Passos, Longbo Huang, Jian Li, and Hang Zhao.
\newblock Lcm-lora: A universal stable-diffusion acceleration module.
\newblock \emph{arXiv preprint arXiv:2311.05556}, 2023{\natexlab{b}}.

\bibitem[Luo(2024)]{luo2024diff}
Weijian Luo.
\newblock Diff-instruct++: Training one-step text-to-image generator model to align with human preferences.
\newblock In \emph{TMLR}, 2024.

\bibitem[Luo et~al.(2023{\natexlab{c}})Luo, Hu, Zhang, Sun, Li, and Zhang]{luo2023diff}
Weijian Luo, Tianyang Hu, Shifeng Zhang, Jiacheng Sun, Zhenguo Li, and Zhihua Zhang.
\newblock Diff-instruct: A universal approach for transferring knowledge from pre-trained diffusion models.
\newblock In \emph{NeurIPS}, pages 76525--76546, 2023{\natexlab{c}}.

\bibitem[Luo et~al.(2024)Luo, Huang, Geng, Kolter, and Qi]{luo2024onestep}
Weijian Luo, Zemin Huang, Zhengyang Geng, J.~Zico Kolter, and Guo-jun Qi.
\newblock One-step diffusion distillation through score implicit matching.
\newblock In \emph{NeurIPS}, 2024.

\bibitem[Luo et~al.(2025{\natexlab{a}})Luo, Zhang, Zhang, and Geng]{luo2024david}
Weijian Luo, Colin Zhang, Debing Zhang, and Zhengyang Geng.
\newblock David and goliath: Small one-step model beats large diffusion with score post-training.
\newblock In \emph{ICML}, 2025{\natexlab{a}}.

\bibitem[Luo et~al.(2025{\natexlab{b}})Luo, Chen, Qu, Hu, and Tang]{luo2024you}
Yihong Luo, Xiaolong Chen, Xinghua Qu, Tianyang Hu, and Jing Tang.
\newblock You only sample once: Taming one-step text-to-image synthesis by self-cooperative diffusion gans.
\newblock In \emph{ICLR}, 2025{\natexlab{b}}.

\bibitem[Ma et~al.(2025{\natexlab{a}})Ma, Wang, Yu, Jia, and Ding]{ma2025msdetreffectivevideomoment}
Hongxu Ma, Guanshuo Wang, Fufu Yu, Qiong Jia, and Shouhong Ding.
\newblock Ms-detr: Towards effective video moment retrieval and highlight detection by joint motion-semantic learning.
\newblock In \emph{ACMMM}, 2025{\natexlab{a}}.

\bibitem[Ma et~al.(2025{\natexlab{b}})Ma, Zhang, Zhang, Zhou, Guan, and Zhou]{ma2025finegrainedzeroshotobjectdetection}
Hongxu Ma, Chenbo Zhang, Lu Zhang, Jiaogen Zhou, Jihong Guan, and Shuigeng Zhou.
\newblock Fine-grained zero-shot object detection.
\newblock In \emph{ACMMM}, 2025{\natexlab{b}}.

\bibitem[Mao et~al.(2024)Mao, Jiang, Wang, Zhu, Zhang, Chen, Chi, and Wang]{mao2024osv}
Xiaofeng Mao, Zhengkai Jiang, Fu-Yun Wang, Wenbing Zhu, Jiangning Zhang, Hao Chen, Mingmin Chi, and Yabiao Wang.
\newblock Osv: One step is enough for high-quality image to video generation.
\newblock \emph{arXiv preprint arXiv:2409.11367}, 2024.

\bibitem[Mi et~al.(2025)Mi, Zhong, Huang, Yuan, Zhao, Xu, Ding, Wang, Guo, and Zhou]{mi2025data}
Yuxi Mi, Zhizhou Zhong, Yuge Huang, Qiuyang Yuan, Xuan Zhao, Jianqing Xu, Shouhong Ding, Shaoming Wang, Rizen Guo, and Shuigeng Zhou.
\newblock Data synthesis with diverse styles for face recognition via 3dmm-guided diffusion.
\newblock In \emph{CVPR}, pages 21203--21214, 2025.

\bibitem[Minka(2005)]{minka2005divergence}
Thomas Minka.
\newblock Divergence measures and message passing.
\newblock \emph{Microsoft Research, Technical Report}, 2005.

\bibitem[{Movie Gen Team}(2024)]{moviegenteam2024movie}
{Movie Gen Team}.
\newblock Movie gen: A cast of media foundation models.
\newblock \url{https://ai.meta.com/static-resource/movie-gen-research-paper}, 2024.

\bibitem[Nan et~al.(2024)Nan, Xie, Zhou, Fan, Yang, Chen, Li, Yang, and Tai]{nan2024openvid1m}
Kepan Nan, Rui Xie, Penghao Zhou, Tiehan Fan, Zhenheng Yang, Zhijie Chen, Xiang Li, Jian Yang, and Ying Tai.
\newblock Openvid-1m: A large-scale high-quality dataset for text-to-video generation.
\newblock \emph{arXiv preprint arXiv:2407.02371}, 2024.

\bibitem[Oquab et~al.(2024)Oquab, Darcet, Moutakanni, Vo, Szafraniec, Khalidov, Fernandez, Haziza, Massa, {El-Nouby}, Assran, Ballas, Galuba, Howes, Huang, Li, Misra, Rabbat, Sharma, Synnaeve, Xu, Jegou, Mairal, Labatut, Joulin, and Bojanowski]{oquab2024dinov2}
Maxime Oquab, Timoth{\'e}e Darcet, Th{\'e}o Moutakanni, Huy Vo, Marc Szafraniec, Vasil Khalidov, Pierre Fernandez, Daniel Haziza, Francisco Massa, Alaaeldin {El-Nouby}, Mahmoud Assran, Nicolas Ballas, Wojciech Galuba, Russell Howes, Po-Yao Huang, Shang-Wen Li, Ishan Misra, Michael Rabbat, Vasu Sharma, Gabriel Synnaeve, Hu Xu, Herv{\'e} Jegou, Julien Mairal, Patrick Labatut, Armand Joulin, and Piotr Bojanowski.
\newblock Dinov2: Learning robust visual features without supervision.
\newblock \emph{TMLR}, 2024.

\bibitem[Peebles and Xie(2023)]{peebles2023scalable}
William Peebles and Saining Xie.
\newblock Scalable diffusion models with transformers.
\newblock In \emph{ICCV}, 2023.

\bibitem[{PKU-Yuan Lab} and {Tuzhan AI}(2024)]{pku-yuanlab2024opensoraplanv110}
{PKU-Yuan Lab} and {Tuzhan AI}.
\newblock Open-sora-plan-v1.1.0.
\newblock \url{https://huggingface.co/datasets/LanguageBind/Open-Sora-Plan-v1.1.0}, 2024.

\bibitem[Poole et~al.(2023)Poole, Jain, Barron, and Mildenhall]{poole2023dreamfusion}
Ben Poole, Ajay Jain, Jonathan~T Barron, and Ben Mildenhall.
\newblock Dreamfusion: Text-to-3d using 2d diffusion.
\newblock In \emph{ICLR}, 2023.

\bibitem[Radford et~al.(2021)Radford, Kim, Hallacy, Ramesh, Goh, Agarwal, Sastry, Askell, Mishkin, Clark, Krueger, and Sutskever]{radford2021learning}
Alec Radford, Jong~Wook Kim, Chris Hallacy, Aditya Ramesh, Gabriel Goh, Sandhini Agarwal, Girish Sastry, Amanda Askell, Pamela Mishkin, Jack Clark, Gretchen Krueger, and Ilya Sutskever.
\newblock Learning transferable visual models from natural language supervision.
\newblock In \emph{ICML}, 2021.

\bibitem[Rajbhandari et~al.(2020)Rajbhandari, Rasley, Ruwase, and He]{rajbhandari2020zero}
Samyam Rajbhandari, Jeff Rasley, Olatunji Ruwase, and Yuxiong He.
\newblock Zero: Memory optimizations toward training trillion parameter models.
\newblock In \emph{SC}, 2020.

\bibitem[Ramachandran et~al.(2017)Ramachandran, Zoph, and Le]{ramachandran2017searching}
Prajit Ramachandran, Barret Zoph, and Quoc~V. Le.
\newblock Searching for activation functions.
\newblock \emph{arXiv preprint arXiv:1710.05941}, 2017.

\bibitem[Ren et~al.(2024)Ren, Xia, Lu, Zhang, Wu, Xie, Wang, and Xiao]{ren2024hypersd}
Yuxi Ren, Xin Xia, Yanzuo Lu, Jiacheng Zhang, Jie Wu, Pan Xie, Xing Wang, and Xuefeng Xiao.
\newblock Hyper-sd: Trajectory segmented consistency model for efficient image synthesis.
\newblock In \emph{NeurIPS}, 2024.

\bibitem[Rombach et~al.(2022)Rombach, Blattmann, Lorenz, Esser, and Ommer]{rombach2022highresolution}
Robin Rombach, Andreas Blattmann, Dominik Lorenz, Patrick Esser, and Bj{\"o}rn Ommer.
\newblock High-resolution image synthesis with latent diffusion models.
\newblock In \emph{CVPR}, 2022.

\bibitem[Ronneberger et~al.(2015)Ronneberger, Fischer, and Brox]{ronneberger2015unet}
Olaf Ronneberger, Philipp Fischer, and Thomas Brox.
\newblock U-net: Convolutional networks for biomedical image segmentation.
\newblock In \emph{MICCAI}, 2015.

\bibitem[Salimans and Ho(2022)]{salimans2022progressive}
Tim Salimans and Jonathan Ho.
\newblock Progressive distillation for fast sampling of diffusion models.
\newblock In \emph{ICLR}, 2022.

\bibitem[Salimans et~al.(2024)Salimans, Mensink, Heek, and Hoogeboom]{salimans2024multistep}
Tim Salimans, Thomas Mensink, Jonathan Heek, and Emiel Hoogeboom.
\newblock Multistep distillation of diffusion models via moment matching.
\newblock In \emph{NeurIPS}, 2024.

\bibitem[Sauer et~al.(2024{\natexlab{a}})Sauer, Boesel, Dockhorn, Blattmann, Esser, and Rombach]{sauer2024fast}
Axel Sauer, Frederic Boesel, Tim Dockhorn, Andreas Blattmann, Patrick Esser, and Robin Rombach.
\newblock Fast high-resolution image synthesis with latent adversarial diffusion distillation.
\newblock \emph{arXiv preprint arXiv:2403.12015}, 2024{\natexlab{a}}.

\bibitem[Sauer et~al.(2024{\natexlab{b}})Sauer, Lorenz, Blattmann, and Rombach]{sauer2023adversarial}
Axel Sauer, Dominik Lorenz, Andreas Blattmann, and Robin Rombach.
\newblock Adversarial diffusion distillation.
\newblock In \emph{ECCV}, 2024{\natexlab{b}}.

\bibitem[Shen et~al.(2023)Shen, Lu, Hu, and Ma]{shen2023collaborative}
Meng Shen, Yanzuo Lu, Yanxu Hu, and Andy~J. Ma.
\newblock Collaborative learning of diverse experts for source-free universal domain adaptation.
\newblock In \emph{ACM MM}, pages 2054--2065, 2023.

\bibitem[Song et~al.(2021{\natexlab{a}})Song, Meng, and Ermon]{song2021denoising}
Jiaming Song, Chenlin Meng, and Stefano Ermon.
\newblock Denoising diffusion implicit models.
\newblock In \emph{ICLR}, 2021{\natexlab{a}}.

\bibitem[Song and Dhariwal(2024)]{song2023improved}
Yang Song and Prafulla Dhariwal.
\newblock Improved techniques for training consistency models.
\newblock In \emph{ICLR}, 2024.

\bibitem[Song et~al.(2021{\natexlab{b}})Song, {Sohl-Dickstein}, Kingma, Kumar, Ermon, and Poole]{song2021scorebased}
Yang Song, Jascha {Sohl-Dickstein}, Diederik~P. Kingma, Abhishek Kumar, Stefano Ermon, and Ben Poole.
\newblock Score-based generative modeling through stochastic differential equations.
\newblock In \emph{ICLR}, 2021{\natexlab{b}}.

\bibitem[Song et~al.(2023)Song, Dhariwal, Chen, and Sutskever]{song2023consistency}
Yang Song, Prafulla Dhariwal, Mark Chen, and Ilya Sutskever.
\newblock Consistency models.
\newblock In \emph{ICML}, 2023.

\bibitem[Sun et~al.(2023)Sun, Pan, Ge, Li, Duan, Wu, Zhang, Zhou, Qin, Wang, Dai, Qiao, Wang, and Li]{sun2023journeydb}
Keqiang Sun, Junting Pan, Yuying Ge, Hao Li, Haodong Duan, Xiaoshi Wu, Renrui Zhang, Aojun Zhou, Zipeng Qin, Yi Wang, Jifeng Dai, Yu Qiao, Limin Wang, and Hongsheng Li.
\newblock Journeydb: A benchmark for generative image understanding.
\newblock \emph{arXiv preprint arXiv:2307.00716}, 2023.

\bibitem[Tan et~al.(2019)Tan, Song, and Ou]{tan2019calibrated}
Zhiqiang Tan, Yunfu Song, and Zhijian Ou.
\newblock Calibrated adversarial algorithms for generative modelling.
\newblock \emph{Stat}, 8:\penalty0 e224, 2019.

\bibitem[Wang et~al.(2024{\natexlab{a}})Wang, Huang, Bergman, Shen, Gao, Lingelbach, Sun, Bian, Song, Liu, Li, and Wang]{wang2024phased}
Fu-Yun Wang, Zhaoyang Huang, Alexander~William Bergman, Dazhong Shen, Peng Gao, Michael Lingelbach, Keqiang Sun, Weikang Bian, Guanglu Song, Yu Liu, Hongsheng Li, and Xiaogang Wang.
\newblock Phased consistency model.
\newblock \emph{arXiv preprint arXiv:2405.18407}, 2024{\natexlab{a}}.

\bibitem[Wang et~al.(2024{\natexlab{b}})Wang, Huang, Bian, Shi, Sun, Song, Liu, and Li]{wang2024animatelcm}
Fu-Yun Wang, Zhaoyang Huang, Weikang Bian, Xiaoyu Shi, Keqiang Sun, Guanglu Song, Yu Liu, and Hongsheng Li.
\newblock Animatelcm: Computation-efficient personalized style video generation without personalized video data.
\newblock In \emph{SIGGRAPH ASIA Technical Communications}, 2024{\natexlab{b}}.

\bibitem[Wang et~al.(2024{\natexlab{c}})Wang, Yang, Huang, Wang, and Li]{wang2024rectified}
Fu-Yun Wang, Ling Yang, Zhaoyang Huang, Mengdi Wang, and Hongsheng Li.
\newblock Rectified diffusion: Straightness is not your need in rectified flow.
\newblock \emph{arXiv preprint arXiv:2410.07303}, 2024{\natexlab{c}}.

\bibitem[Wang and Kanwar(2019)]{wang2019bfloat16}
Shibo Wang and Pankaj Kanwar.
\newblock Bfloat16: The secret to high performance on cloud tpus.
\newblock \url{https://cloud.google.com/blog/products/ai-machine-learning/bfloat16-the-secret-to-high-performance-on-cloud-tpus}, 2019.

\bibitem[Wang et~al.(2024{\natexlab{d}})Wang, Zhang, Luo, Sun, Cui, Wang, Zhang, Wang, Li, Yu, Zhao, Ao, Min, Li, Wu, Zhao, Zhang, Wang, Liu, He, Yang, Liu, Lin, Huang, and Wang]{wang2024emu3}
Xinlong Wang, Xiaosong Zhang, Zhengxiong Luo, Quan Sun, Yufeng Cui, Jinsheng Wang, Fan Zhang, Yueze Wang, Zhen Li, Qiying Yu, Yingli Zhao, Yulong Ao, Xuebin Min, Tao Li, Boya Wu, Bo Zhao, Bowen Zhang, Liangdong Wang, Guang Liu, Zheqi He, Xi Yang, Jingjing Liu, Yonghua Lin, Tiejun Huang, and Zhongyuan Wang.
\newblock Emu3: Next-token prediction is all you need.
\newblock \emph{arXiv preprint arXiv:2409.18869}, 2024{\natexlab{d}}.

\bibitem[Wang et~al.(2023)Wang, Lu, Wang, Bao, Li, Su, and Zhu]{wang2023prolificdreamer}
Zhengyi Wang, Cheng Lu, Yikai Wang, Fan Bao, Chongxuan Li, Hang Su, and Jun Zhu.
\newblock Prolificdreamer: High-fidelity and diverse text-to-3d generation with variational score distillation.
\newblock In \emph{NeurIPS}, 2023.

\bibitem[Wu et~al.(2023)Wu, Hao, Sun, Chen, Zhu, Zhao, and Li]{wu2023human}
Xiaoshi Wu, Yiming Hao, Keqiang Sun, Yixiong Chen, Feng Zhu, Rui Zhao, and Hongsheng Li.
\newblock Human preference score v2: A solid benchmark for evaluating human preferences of text-to-image synthesis.
\newblock \emph{arXiv preprint arXiv:2306.09341}, 2023.

\bibitem[Wu and He(2018)]{wu2018group}
Yuxin Wu and Kaiming He.
\newblock Group normalization.
\newblock In \emph{ECCV}, 2018.

\bibitem[Xiao et~al.(2017)Xiao, Jin, Yang, Yang, Sun, and Chang]{xiao2017building}
Xuefeng Xiao, Lianwen Jin, Yafeng Yang, Weixin Yang, Jun Sun, and Tianhai Chang.
\newblock Building fast and compact convolutional neural networks for offline handwritten chinese character recognition.
\newblock \emph{Pattern Recognition}, 72:\penalty0 72--81, 2017.

\bibitem[Xu et~al.(2023)Xu, Liu, Wu, Tong, Li, Ding, Tang, and Dong]{xu2023imagereward}
Jiazheng Xu, Xiao Liu, Yuchen Wu, Yuxuan Tong, Qinkai Li, Ming Ding, Jie Tang, and Yuxiao Dong.
\newblock Imagereward: Learning and evaluating human preferences for text-to-image generation.
\newblock In \emph{NeurIPS}, 2023.

\bibitem[Xu et~al.(2024)Xu, Zhao, Xiao, and Hou]{xu2024ufogen}
Yanwu Xu, Yang Zhao, Zhisheng Xiao, and Tingbo Hou.
\newblock Ufogen: You forward once large scale text-to-image generation via diffusion gans.
\newblock In \emph{CVPR}, 2024.

\bibitem[Xu et~al.(2025)Xu, Yu, Zhou, Zhou, Jin, Hong, Ji, Zhu, Cai, Tang, et~al.]{xu2025hunyuanportrait}
Zunnan Xu, Zhentao Yu, Zixiang Zhou, Jun Zhou, Xiaoyu Jin, Fa-Ting Hong, Xiaozhong Ji, Junwei Zhu, Chengfei Cai, Shiyu Tang, et~al.
\newblock Hunyuanportrait: Implicit condition control for enhanced portrait animation.
\newblock In \emph{CVPR}, pages 15909--15919, 2025.

\bibitem[Yan et~al.(2024)Yan, Liu, Pan, Liew, Liu, and Feng]{yan2024perflow}
Hanshu Yan, Xingchao Liu, Jiachun Pan, Jun~Hao Liew, Qiang Liu, and Jiashi Feng.
\newblock Perflow: Piecewise rectified flow as universal plug-and-play accelerator.
\newblock In \emph{NeurIPS}, 2024.

\bibitem[Yang et~al.(2024{\natexlab{a}})Yang, Huang, Lu, Han, Zhang, Gao, Hu, and Zhao]{yang2024vript}
Dongjie Yang, Suyuan Huang, Chengqiang Lu, Xiaodong Han, Haoxin Zhang, Yan Gao, Yao Hu, and Hai Zhao.
\newblock Vript: A video is worth thousands of words.
\newblock \emph{arXiv preprint arXiv:2406.06040}, 2024{\natexlab{a}}.

\bibitem[Yang et~al.(2024{\natexlab{b}})Yang, Teng, Zheng, Ding, Huang, Xu, Yang, Hong, Zhang, Feng, Yin, Gu, Zhang, Wang, Cheng, Liu, Xu, Dong, and Tang]{yang2024cogvideox}
Zhuoyi Yang, Jiayan Teng, Wendi Zheng, Ming Ding, Shiyu Huang, Jiazheng Xu, Yuanming Yang, Wenyi Hong, Xiaohan Zhang, Guanyu Feng, Da Yin, Xiaotao Gu, Yuxuan Zhang, Weihan Wang, Yean Cheng, Ting Liu, Bin Xu, Yuxiao Dong, and Jie Tang.
\newblock Cogvideox: Text-to-video diffusion models with an expert transformer.
\newblock \emph{arXiv preprint arXiv:2408.06072}, 2024{\natexlab{b}}.

\bibitem[Yi et~al.(2025)Yi, Shao, Ye, Zhao, Yin, Lingelbach, Yuan, Tian, Xie, and Zhou]{yi2025magic}
Hongwei Yi, Shitong Shao, Tian Ye, Jiantong Zhao, Qingyu Yin, Michael Lingelbach, Li Yuan, Yonghong Tian, Enze Xie, and Daquan Zhou.
\newblock Magic 1-for-1: Generating one minute video clips within one minute.
\newblock \emph{arXiv preprint arXiv:2502.07701}, 2025.

\bibitem[Yin et~al.(2024{\natexlab{a}})Yin, Gharbi, Park, Zhang, Shechtman, Durand, and Freeman]{yin2024improved}
Tianwei Yin, Micha{\"e}l Gharbi, Taesung Park, Richard Zhang, Eli Shechtman, Fredo Durand, and William~T. Freeman.
\newblock Improved distribution matching distillation for fast image synthesis.
\newblock In \emph{NeurIPS}, 2024{\natexlab{a}}.

\bibitem[Yin et~al.(2024{\natexlab{b}})Yin, Gharbi, Zhang, Shechtman, Durand, Freeman, and Park]{yin2024onestep}
Tianwei Yin, Micha{\"e}l Gharbi, Richard Zhang, Eli Shechtman, Fr{\'e}do Durand, William~T. Freeman, and Taesung Park.
\newblock One-step diffusion with distribution matching distillation.
\newblock In \emph{CVPR}, pages 6613--6623, 2024{\natexlab{b}}.

\bibitem[Yin et~al.(2024{\natexlab{c}})Yin, Zhang, Zhang, Freeman, Durand, Shechtman, and Huang]{yin2024slow}
Tianwei Yin, Qiang Zhang, Richard Zhang, William~T Freeman, Fredo Durand, Eli Shechtman, and Xun Huang.
\newblock From slow bidirectional to fast causal video generators.
\newblock \emph{arXiv preprint arXiv:2412.07772}, 2024{\natexlab{c}}.

\bibitem[Yu et~al.(2022)Yu, Xu, Koh, Luong, Baid, Wang, Vasudevan, Ku, Yang, Ayan, Hutchinson, Han, Parekh, Li, Zhang, Baldridge, and Wu]{yu2022scaling}
Jiahui Yu, Yuanzhong Xu, Jing~Yu Koh, Thang Luong, Gunjan Baid, Zirui Wang, Vijay Vasudevan, Alexander Ku, Yinfei Yang, Burcu~Karagol Ayan, Ben Hutchinson, Wei Han, Zarana Parekh, Xin Li, Han Zhang, Jason Baldridge, and Yonghui Wu.
\newblock Scaling autoregressive models for content-rich text-to-image generation.
\newblock \emph{TMLR}, 2022.

\bibitem[Zhang et~al.(2018)Zhang, Isola, Efros, Shechtman, and Wang]{zhang2018unreasonable}
Richard Zhang, Phillip Isola, Alexei~A. Efros, Eli Shechtman, and Oliver Wang.
\newblock The unreasonable effectiveness of deep features as a perceptual metric.
\newblock In \emph{CVPR}, pages 586--595, 2018.

\bibitem[Zhang et~al.(2024{\natexlab{a}})Zhang, Wang, Wu, Li, Gao, Zhang, and Wang]{zhang2024learning}
Sixian Zhang, Bohan Wang, Junqiang Wu, Yan Li, Tingting Gao, Di Zhang, and Zhongyuan Wang.
\newblock Learning multi-dimensional human preference for text-to-image generation.
\newblock In \emph{CVPR}, pages 8018--8027, 2024{\natexlab{a}}.

\bibitem[Zhang et~al.(2024{\natexlab{b}})Zhang, Li, Wu, Xu, Kag, Skorokhodov, Menapace, Siarohin, Cao, Metaxas, Tulyakov, and Ren]{zhang2024sfv}
Zhixing Zhang, Yanyu Li, Yushu Wu, Yanwu Xu, Anil Kag, Ivan Skorokhodov, Willi Menapace, Aliaksandr Siarohin, Junli Cao, Dimitris Metaxas, Sergey Tulyakov, and Jian Ren.
\newblock Sf-v: Single forward video generation model.
\newblock In \emph{NeurIPS}, 2024{\natexlab{b}}.

\bibitem[Zheng et~al.(2024)Zheng, Hu, Fan, Wang, Ding, Tao, and Cham]{zheng2024trajectory}
Jianbin Zheng, Minghui Hu, Zhongyi Fan, Chaoyue Wang, Changxing Ding, Dacheng Tao, and Tat-Jen Cham.
\newblock Trajectory consistency distillation: Improved latent consistency distillation by semi-linear consistency function with trajectory mapping.
\newblock \emph{arXiv preprint arXiv:2402.19159}, 2024.

\bibitem[Zhou et~al.(2024)Zhou, Zheng, Wang, Yin, and Huang]{zhou2024score}
Mingyuan Zhou, Huangjie Zheng, Zhendong Wang, Mingzhang Yin, and Hai Huang.
\newblock Score identity distillation: Exponentially fast distillation of pretrained diffusion models for one-step generation.
\newblock In \emph{ICML}, 2024.

\bibitem[Zhou et~al.(2025)Zhou, Zheng, Gu, Wang, and Huang]{zhou2024adversarial}
Mingyuan Zhou, Huangjie Zheng, Yi Gu, Zhendong Wang, and Hai Huang.
\newblock Adversarial score identity distillation: Rapidly surpassing the teacher in one step.
\newblock In \emph{ICLR}, 2025.

\end{thebibliography}
}

\clearpage
\setcounter{page}{1}
\maketitlesupplementary

\appendix

\section{Adversarial Distribution Matching}\label{app:algorithm}

During the ADM distillation process, the fake score estimator, generator, and discriminator are updated alternately. 
The \cref{alg:dmdx} below clarifies the training procedure.
Our ablation experiments in \cref{sec:ablation} demonstrate that TTUR has minimal impact on the final performance. 
Therefore, in our experiments, we set TTUR to 1, meaning that the fake model and generator are updated at the same frequency.

\begin{algorithm}[H]
    \footnotesize
    \caption{ADM Training Procedure}
    \label{alg:dmdx}
    \begin{algorithmic}[1]
        \State \textbf{Input:} pretrained teacher model as real score estimator $\bm{F}_\phi$
        \State \textbf{Output:} few-step generator $\bm{G}_\theta$ with schedule $\{t_0,t_1,...,t_N\}$
        \State \textbf{Initialize: } fake score estimator $\bm{f}_\psi\leftarrow\bm{F}_\phi$, generator $\bm{G}_\theta\leftarrow\bm{F}_\phi$, latent-space discriminator $\bm{D}_\tau\leftarrow\bm{F}_\phi$ with multiple trainable heads, generator iteration $genIter\leftarrow 0$, global iteration $globalIter\leftarrow 0$
        \While{$genIter < maxIter$}
            \State $globalIter \mathrel{+}= 1$

            \State 
            \State $//\ update\ fake\ score\ estimator\ \bm{f}_\psi$
            \State sample pure noise $\bm{z}\sim\mathcal{N}(\bm{0},\bm{I})$
            \State solve the PF-ODE w.r.t. $N$ steps in schedule $\bm{x}_0\leftarrow\bm{G}_\theta(\bm{z},\cdot)$
            \State sample new pure noise $\bm{z}_f$ and random timestep $t_f$  
            \State update $\psi$ with $(\bm{x}_0,t_f,\bm{z}_f)$ and pretrain loss in \cref{eq:ddim_pretrain_loss} or \cref{eq:flowmatch_pretrain_loss}
            \State \textbf{if not } $(globalIter\ \%\ \text{TTUR}) == 0$ \textbf{ then continue}
            
            \State 
            \State $//\ update\ generator\ \bm{G}_\theta$
            \State sample pure noise $\hat{\bm{z}}$ and random index $n\in[1,N]$
            \State solve the PF-ODE w/o grad following $t_N \rightarrow t_{N-1} \rightarrow ... \rightarrow t_{n}$, i.e. $\hat{\bm{z}}=\hat{\bm{x}}_{t_{N}}\rightarrow \hat{\bm{x}}_{t_{N-1}}\rightarrow ... \rightarrow \hat{\bm{x}}_{t_{n}}$
            \State solve the PF-ODE w/ grad w.r.t. $t_0$, i.e. $\hat{\bm{x}}_0=\bm{G}_\theta(\hat{\bm{x}}_{t_{n}}, t_{n})$
            \State sample new pure noise $\bm{z}_g$ and random timestep $t\sim\mathcal{U}(0,T)$
            \State diffuse sample $\hat{\bm{x}}_0$ with $\bm{z}_g$ and \cref{eq:diffusion}, i.e. $\bm{x}_{t}=\bm{q}(\bm{x}_{t}|\hat{\bm{x}}_0)$
            \State solve the PF-ODE of $\bm{f}_\psi$ w.r.t. $(t-\Delta t)$ to obtain ${\bm{x}}_{t-\Delta t}^{\text{fake}}$
            \State solve the PF-ODE of $\bm{F}_\phi$ w.r.t. $(t-\Delta t)$ to obtain ${\bm{x}}_{t-\Delta t}^{\text{real}}$
            \State update $\theta$ with $({\bm{x}}_{t-\Delta t}^{\text{fake}},t-\Delta t)$ and \cref{eq:gan_generator_loss}
            \State $genIter \mathrel{+}= 1$
            
            \State 
            \State $//\ update\ discriminator\ \bm{D}_\tau$
            \State update $\tau$ with $({\bm{x}}_{t-\Delta t}^{\text{fake}},{\bm{x}}_{t-\Delta t}^{\text{real}},t-\Delta t)$ and \cref{eq:gan_discriminator_loss}
        \EndWhile
    \end{algorithmic}
\end{algorithm}

\section{Implementation Details}\label{appendix:implementation_details}

\begin{figure*}[t]
\centering
\includegraphics[width=1.0\linewidth]{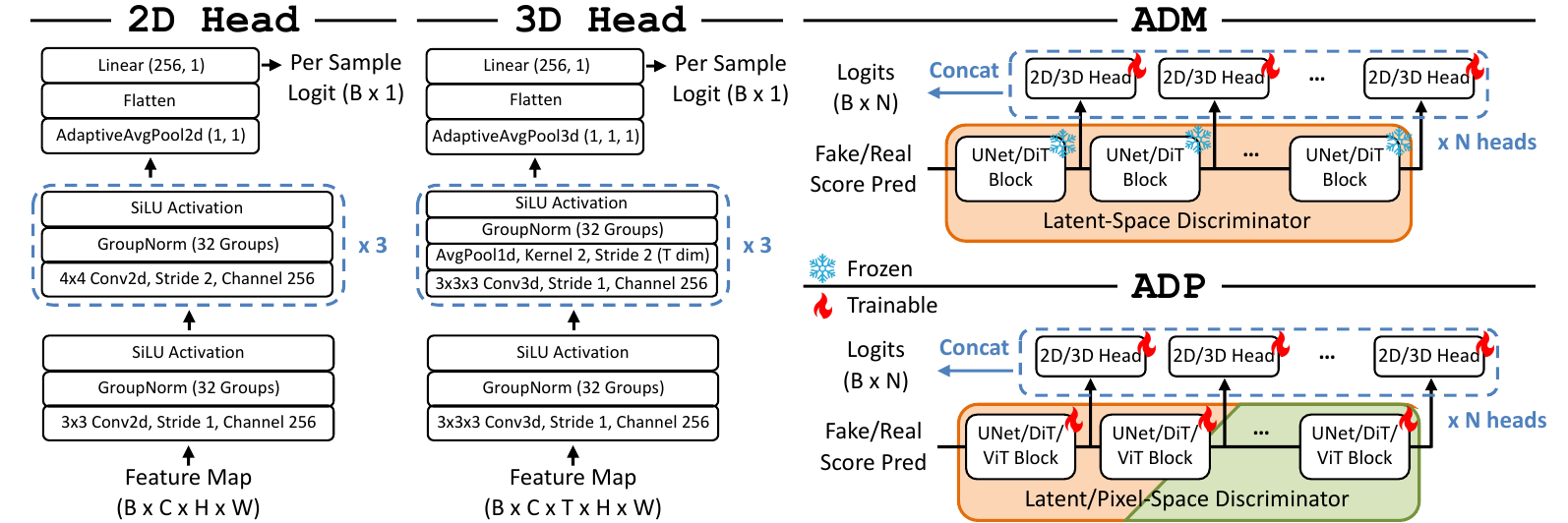}
\caption{Illustration of our discriminator design and the difference between ADM and ADP.}
\label{fig:disc_design}
\end{figure*}
 
\subsection{2D Discriminator Design}
In \cref{fig:disc_design}, we thoroughly illustrate the design of our discriminators and the difference between two training stages.
For all the trainable heads appended to discriminator backbone for text-to-image experiments, we have a fixed 2D design following SDXL-Lightning~\cite{lin2024sdxllightning}, which consists of simple blocks of 4$\times$4 2D convolution with a stride of 2, group normalization~\cite{wu2018group} with 32 groups, and SiLU activation~\cite{hendrycks2016gaussian,ramachandran2017searching} layer.
The difference is that we will append multiple heads at different layers of the network.
Whether it is the output of UNet~\cite{ronneberger2015unet}, DiT~\cite{peebles2023scalable} or ViT~\cite{dosovitskiy2021image}, we uniformly reshape it into $[\bm{B}atch, \bm{C}hannel, \bm{H}eight, \bm{W}idth]$ and then use it as the input to the discriminator head.
For {SDXL~\cite{rombach2022highresolution}}, we take the output of the last ResNet~\cite{he2016deep} of each block (including down-sampling, mid and up-sampling blocks), yielding a total of 7 discriminator heads.
For {SD3 series~\cite{esser2024scaling}} models, we take the output of each DiT block, yielding 24 and 38 discriminator heads for SD3-Medium and SD3.5-Large, respectively.
For {SAM~\cite{kirillov2023segment} and DINOv2~\cite{oquab2024dinov2}}, we take the output of layers 3, 6, 9 and 12, yielding 4 discriminator heads.

\subsection{3D Discriminator Design}
Our 3D discriminator head for text-to-video latent diffusion models consists of simple blocks of 3$\times$3$\times$3 3D convolution with a stride of 1, 3$\times$3 2D convolution with a stride of 2, group normalization with 32 groups and SiLU activation layer.
This is similar to the design in 2D discriminator head except that we additionally insert several 3D convolution layers to extract time-dependent feature.
The output of specific blocks within video DiT backbone are reshaped into $[\bm{B}atch, \bm{C}hannel, \bm{T}ime, \bm{H}eight, \bm{W}idth]$ and input to corresponding discriminator head.
In practice, we extract features every 3 DiT blocks due to the computational effort of 3D convolution, yielding a total of 10 and 14 discrimiantor heads for 2B and 5B models, respectively.



\begin{table}[t]
    \centering
    \footnotesize
    \begin{tabular}{*{7}{m{0.75cm}}}
        \toprule[1.5pt]

         & 
         \makebox[0.75cm][c]{\makecell[c]{\textbf{Training}\\\textbf{Iteration}}} & 
         \makebox[0.75cm][c]{\makecell[c]{\textbf{GPU}\\\textbf{Number}}} & 
         \makebox[0.75cm][c]{\makecell[c]{\textbf{Elapsed}\\\textbf{Time}}} & 
         \makebox[0.75cm][c]{\makecell[c]{\textbf{GPU}\\\textbf{Hours}}} & 
         \makebox[0.75cm][c]{\makecell[c]{\textbf{Micro}\\\textbf{BatchSize}}} &
         \makebox[0.75cm][c]{\makecell[c]{\textbf{Max}\\\textbf{Memory}}} \\
        
        \midrule[1.5pt]

        \makebox[0.75cm][l]{DMD2} & \makebox[0.75cm][c]{20K} & \makebox[0.75cm][c]{64} & \makebox[0.75cm][c]{60 hours} & \makebox[0.75cm][c]{3840} & \makebox[0.75cm][c]{2} & \makebox[0.75cm][c]{-} \\

        \makebox[0.75cm][l]{\textbf{DMDX}} & \makebox[0.75cm][c]{\textbf{8K+8K}} & \makebox[0.75cm][c]{\textbf{32}} & \makebox[0.75cm][c]{\textbf{70 hours}} & \makebox[0.75cm][c]{\textbf{2240}} & \makebox[0.75cm][c]{\textbf{4}} & \makebox[0.75cm][c]{\textbf{39.6 GiB}} \\

        \makebox[0.75cm][l]{- ADP} & \makebox[0.75cm][c]{8K} & \makebox[0.75cm][c]{32} & \makebox[0.75cm][c]{55 hours} & \makebox[0.75cm][c]{1760} & \makebox[0.75cm][c]{4} & \makebox[0.75cm][c]{39.6 GiB} \\

        \makebox[0.75cm][l]{- ADM} & \makebox[0.75cm][c]{8K} & \makebox[0.75cm][c]{32} & \makebox[0.75cm][c]{15 hours} & \makebox[0.75cm][c]{480} & \makebox[0.75cm][c]{4} & \makebox[0.75cm][c]{24.1 GiB} \\

        \bottomrule[1.5pt]
    \end{tabular}
    \vspace{-0.2cm}
    \caption{
        Comparisons on A100 GPU efficiency with DMD2. The elapsed time for ADP already includes collection of ODE pairs.
    }
    \vspace{-0.2cm}
    \label{tab:ablation_efficiency}
\end{table}

\subsection{GPU efficiency.}
In \cref{tab:ablation_efficiency}, we present the training configurations and GPU consumption of our proposed method compared to DMD2.
The table demonstrates that we actually achieve better performance over DMD2 with less GPU time and don't impose excessive demands on GPU memory.
Although maintaining more networks during training process, our implementation attains manageable memory footprint with several optimizations detailed later.

\subsection{Memory efficiency.}
To reduce GPU memory footprint and improve efficiency, we utilize several acceleration techniques in our implementation including {Fully Sharded Data Parallel (FSDP)}~\cite{rajbhandari2020zero}, gradient checkpoint~\cite{chen2016training} and BF16 mixed precision~\cite{wang2019bfloat16}.
For text-to-video models, we additionally integrate, {Context Parallel (CP)}~\cite{yang2024cogvideox} and {Sequence Parallel (SP)}~\cite{jacobs2023deepspeed} following common practice in MovieGen\cite{moviegenteam2024movie} to speed up training and inference.
More importantly, a CPU offloading technique that has been built into Pytorch FSDP is essential for training multiple networks to save memory.

With CPU offloading enabled, each parameter along with the corresponding gradient and optimizer state can be offloaded from the GPU to CPU memory.
In conjunction with gradient checkpointing, the GPU memory footprint in the forward and backward process is nearly the same as when there is only one single network, because the peak memory is now determined by the maximum activation of each block.
This comes at the cost of increased CPU memory and longer time per iteration.
While the CPU memory is usually sufficient and cheap, our more effective approaches require fewer iterations to achieve convergence and satisfactory results, and as \cref{tab:ablation_efficiency} show that our DMDX takes less time than DMD2 on one-step SDXL distillation.

\subsection{Hyperparameters.}
For all models of the optimizer (including generator, fake model and discriminator in both text-to-image and text-to-video experiments), we use AdamW~\cite{loshchilov2019decoupled} optimizer without weight decay, with beta parameters (0.0, 0.99) to capture the changes in distribution more up-to-date.
The learning rates of discriminator and fake model across all of our experiments are fixed at 5e-6 and 1e-6, respectively.

For SDXL, the learning rates for generator during ADP and ADM training are 1e-6 and 1e-7, respectively.
As for multi-step ADM distillation, the learning rates for generator of SD3-Medium LoRA training and SD3.5-Large fully fine-tuning are given to 1e-6 and 1e-8, respectively.
In case of text-to-video diffusion distillation, we set the same learning rate 1e-7 for different 8-step CogVideoX generators.

Among all the ADM experiments, the Classifier-Free Guidance (CFG) is required for real model as DMD does~\cite{yin2024improved}.
For SDXL, SD3-Medium, SD3.5-Large, and CogVideoX, the uniformly random sampling ranges for the CFG values are set to [6.0, 8.0], [6.0, 8.0], [3.0, 4.0], and [5.0, 7.0], respectively. 
The chosen ranges are based on the recommended CFG values from the original baseline's inference with some allowable variations.
We observed that this setting is adequate for achieving satisfactory distilled performance without requiring extensive tuning.

The fake model training does not incorporate CFG and uses the same loss function as the standard pre-training of diffusion models, except that we didn't set any dropout. 
For noise-parameterized models, the prediction target is noise, while for velocity-parameterized models, it is velocity.

\section{Main Results}\label{appendix:main_results}

\begin{table*}[t]
    \centering
    \footnotesize
    \begin{tabular}{m{2.7cm}m{0.3cm}m{0.3cm}*{3}{m{0.7cm}}*{4}{m{1.1cm}}*{3}{m{0.7cm}}}
        \toprule[1.5pt]
        \multirow{1}[2]{*}{\makebox[2.7cm][l]{\textbf{Method}}} & 
        \multirow{1}[2]{*}{\makebox[0.3cm][c]{\textbf{Step}}} &
        \multirow{1}[2]{*}{\makebox[0.3cm][c]{\textbf{NFE}}} &
        \makebox[0.7cm][c]{\makecell[c]{\textbf{Final}\\\textbf{Score}}} & 
        \makebox[0.7cm][c]{\makecell[c]{\textbf{Quality}\\\textbf{Score}}} & 
        \makebox[0.7cm][c]{\makecell[c]{\textbf{Semantic}\\\textbf{Score}}} & 
        \makebox[1.1cm][c]{\makecell[c]{Subject\\Consistency}} & 
        \makebox[1.1cm][c]{\makecell[c]{Background\\Consistency}} & 
        \makebox[1.1cm][c]{\makecell[c]{Temporal\\Flickering}} & 
        \makebox[1.1cm][c]{\makecell[c]{Motion\\Smoothness}} & 
        \makebox[0.7cm][c]{\makecell[c]{Dynamic\\Degree}} & 
        \makebox[0.7cm][c]{\makecell[c]{Aesthetic\\Quality}} & 
        \makebox[0.7cm][c]{\makecell[c]{Imaging\\Quality}} \\

        \midrule[1.5pt]

        \makebox[2.7cm][l]{\textbf{ADM}} & 
        \makebox[0.3cm][c]{8} &
        \makebox[0.3cm][c]{8} &
        \makebox[0.7cm][c]{78.58} &
        \makebox[0.7cm][c]{80.82} &
        \makebox[0.7cm][c]{69.62} &
        \makebox[1.1cm][c]{96.72} &
        \makebox[1.1cm][c]{96.55} &
        \makebox[1.1cm][c]{97.01} &
        \makebox[1.1cm][c]{98.14} &
        \makebox[0.7cm][c]{48.61} &
        \makebox[0.7cm][c]{57.80} &
        \makebox[0.7cm][c]{65.28} \\

        \makebox[2.7cm][l]{+Longer Training $\times$2} & 
        \makebox[0.3cm][c]{8} &
        \makebox[0.3cm][c]{8} &
        \makebox[0.7cm][c]{80.76} &
        \makebox[0.7cm][c]{\textbf{83.03}} &
        \makebox[0.7cm][c]{71.69} &
        \makebox[1.1cm][c]{96.58} &
        \makebox[1.1cm][c]{96.71} &
        \makebox[1.1cm][c]{98.12} &
        \makebox[1.1cm][c]{97.68} &
        \makebox[0.7cm][c]{73.33} &
        \makebox[0.7cm][c]{57.90} &
        \makebox[0.7cm][c]{65.72} \\

        \makebox[2.7cm][l]{\textbf{ADM w/ CFG}} & 
        \makebox[0.3cm][c]{8} &
        \makebox[0.3cm][c]{16} &
        \makebox[0.7cm][c]{79.86} &
        \makebox[0.7cm][c]{80.93} &
        \makebox[0.7cm][c]{75.56} &
        \makebox[1.1cm][c]{96.16} &
        \makebox[1.1cm][c]{96.96} &
        \makebox[1.1cm][c]{96.86} &
        \makebox[1.1cm][c]{97.69} &
        \makebox[0.7cm][c]{54.44} &
        \makebox[0.7cm][c]{59.78} &
        \makebox[0.7cm][c]{63.18} \\

        \makebox[2.7cm][l]{+Longer Training $\times$2} & 
        \makebox[0.3cm][c]{8} &
        \makebox[0.3cm][c]{16} &
        \makebox[0.7cm][c]{\textbf{81.79}} &
        \makebox[0.7cm][c]{83.00} &
        \makebox[0.7cm][c]{\textbf{76.94}} &
        \makebox[1.1cm][c]{96.83} &
        \makebox[1.1cm][c]{96.90} &
        \makebox[1.1cm][c]{98.51} &
        \makebox[1.1cm][c]{98.07} &
        \makebox[0.7cm][c]{63.05} &
        \makebox[0.7cm][c]{61.03} &
        \makebox[0.7cm][c]{64.62} \\

        \makebox[2.7cm][l]{\textcolor{gray}{CogVideoX-2b}} & 
        \makebox[0.3cm][c]{\textcolor{gray}{100}} &
        \makebox[0.3cm][c]{\textcolor{gray}{200}} &
        \makebox[0.7cm][c]{\textcolor{gray}{80.03}} &
        \makebox[0.7cm][c]{\textcolor{gray}{80.80}} &
        \makebox[0.7cm][c]{\textcolor{gray}{76.97}} &
        \makebox[1.1cm][c]{\textcolor{gray}{92.53}} &
        \makebox[1.1cm][c]{\textcolor{gray}{95.22}} &
        \makebox[1.1cm][c]{\textcolor{gray}{97.79}} &
        \makebox[1.1cm][c]{\textcolor{gray}{97.00}} &
        \makebox[0.7cm][c]{\textcolor{gray}{69.44}} &
        \makebox[0.7cm][c]{\textcolor{gray}{60.38}} &
        \makebox[0.7cm][c]{\textcolor{gray}{60.69}} \\

        \midrule[0.75pt]

        \makebox[2.7cm][l]{\textbf{ADM}} & 
        \makebox[0.3cm][c]{8} &
        \makebox[0.3cm][c]{8} &
        \makebox[0.7cm][c]{\textbf{82.06}} &
        \makebox[0.7cm][c]{\textbf{83.22}} &
        \makebox[0.7cm][c]{\textbf{77.42}} &
        \makebox[1.1cm][c]{96.42} &
        \makebox[1.1cm][c]{96.87} &
        \makebox[1.1cm][c]{96.96} &
        \makebox[1.1cm][c]{97.69} &
        \makebox[0.7cm][c]{68.88} &
        \makebox[0.7cm][c]{61.17} &
        \makebox[0.7cm][c]{69.01} \\


        \makebox[2.7cm][l]{\textbf{ADM w/ CFG}} & 
        \makebox[0.3cm][c]{8} &
        \makebox[0.3cm][c]{16} &
        \makebox[0.7cm][c]{80.98} &
        \makebox[0.7cm][c]{82.16} &
        \makebox[0.7cm][c]{76.25} &
        \makebox[1.1cm][c]{96.15} &
        \makebox[1.1cm][c]{96.59} &
        \makebox[1.1cm][c]{95.99} &
        \makebox[1.1cm][c]{98.57} &
        \makebox[0.7cm][c]{56.66} &
        \makebox[0.7cm][c]{61.01} &
        \makebox[0.7cm][c]{68.68} \\


        \makebox[2.7cm][l]{\textcolor{gray}{CogVideoX-5b}} & 
        \makebox[0.3cm][c]{\textcolor{gray}{100}} &
        \makebox[0.3cm][c]{\textcolor{gray}{200}} &
        \makebox[0.7cm][c]{\textcolor{gray}{81.22}} &
        \makebox[0.7cm][c]{\textcolor{gray}{81.78}} &
        \makebox[0.7cm][c]{\textcolor{gray}{78.98}} &
        \makebox[1.1cm][c]{\textcolor{gray}{92.52}} &
        \makebox[1.1cm][c]{\textcolor{gray}{96.68}} &
        \makebox[1.1cm][c]{\textcolor{gray}{98.34}} &
        \makebox[1.1cm][c]{\textcolor{gray}{96.97}} &
        \makebox[0.7cm][c]{\textcolor{gray}{70.55}} &
        \makebox[0.7cm][c]{\textcolor{gray}{61.67}} &
        \makebox[0.7cm][c]{\textcolor{gray}{61.88}} \\

        \bottomrule[1.5pt]
    \end{tabular}
    \caption{
        VBench~\cite{huang2024vbench} detailed results on \textbf{overall scores} and separate score for each quality dimension.
    }
    \label{tab:vbench_1}

    \vspace*{0.3cm}

    \begin{tabular}{m{2.7cm}m{0.3cm}m{0.3cm}*{3}{m{0.9cm}}m{0.7cm}m{1.3cm}m{0.7cm}*{3}{m{1.2cm}}}
        \toprule[1.5pt]
        \multirow{1}[2]{*}{\makebox[2.7cm][l]{\textbf{Method}}} & 
        \multirow{1}[2]{*}{\makebox[0.3cm][c]{\textbf{Step}}} &
        \multirow{1}[2]{*}{\makebox[0.3cm][c]{\textbf{NFE}}} &
        \makebox[0.9cm][c]{\makecell[c]{Object\\Class}} & 
        \makebox[0.9cm][c]{\makecell[c]{Multiple\\Objects}} & 
        \makebox[0.9cm][c]{\makecell[c]{Human\\Action}} & 
        \multirow{1}[2]{*}{\makebox[0.7cm][c]{\makecell[c]{Color}}} & 
        \makebox[1.3cm][c]{\makecell[c]{Spatial\\Relationship}} & 
        \multirow{1}[2]{*}{\makebox[0.7cm][c]{\makecell[c]{Scene}}} & 
        \makebox[1.2cm][c]{\makecell[c]{Appearance\\Style}} & 
        \makebox[1.2cm][c]{\makecell[c]{Temporal\\Style}} & 
        \makebox[1.2cm][c]{\makecell[c]{Overall\\Consistency}} \\
        
        \midrule[1.5pt]

        \makebox[2.7cm][l]{\textbf{ADM}} & 
        \makebox[0.3cm][c]{8} &
        \makebox[0.3cm][c]{8} &
        \makebox[0.9cm][c]{83.97} &
        \makebox[0.9cm][c]{47.19} &
        \makebox[0.9cm][c]{87.40} &
        \makebox[0.7cm][c]{77.79} &
        \makebox[1.3cm][c]{62.93} &
        \makebox[0.7cm][c]{42.64} &
        \makebox[1.2cm][c]{24.16} &
        \makebox[1.2cm][c]{22.35} &
        \makebox[1.2cm][c]{25.27} \\

        \makebox[2.7cm][l]{+Longer Training $\times$2} & 
        \makebox[0.3cm][c]{8} &
        \makebox[0.3cm][c]{8} &
        \makebox[0.9cm][c]{87.84} &
        \makebox[0.9cm][c]{56.53} &
        \makebox[0.9cm][c]{85.00} &
        \makebox[0.7cm][c]{80.28} &
        \makebox[1.3cm][c]{69.52} &
        \makebox[0.7cm][c]{44.33} &
        \makebox[1.2cm][c]{23.15} &
        \makebox[1.2cm][c]{22.60} &
        \makebox[1.2cm][c]{25.11} \\

        \makebox[2.7cm][l]{\textbf{ADM w/ CFG}} & 
        \makebox[0.3cm][c]{8} &
        \makebox[0.3cm][c]{16} &
        \makebox[0.9cm][c]{89.55} &
        \makebox[0.9cm][c]{64.78} &
        \makebox[0.9cm][c]{92.60} &
        \makebox[0.7cm][c]{82.31} &
        \makebox[1.3cm][c]{62.61} &
        \makebox[0.7cm][c]{52.73} &
        \makebox[1.2cm][c]{24.31} &
        \makebox[1.2cm][c]{24.46} &
        \makebox[1.2cm][c]{26.12} \\

        \makebox[2.7cm][l]{+Longer Training $\times$2} & 
        \makebox[0.3cm][c]{8} &
        \makebox[0.3cm][c]{16} &
        \makebox[0.9cm][c]{91.67} &
        \makebox[0.9cm][c]{71.58} &
        \makebox[0.9cm][c]{92.20} &
        \makebox[0.7cm][c]{82.01} &
        \makebox[1.3cm][c]{71.79} &
        \makebox[0.7cm][c]{50.26} &
        \makebox[1.2cm][c]{23.54} &
        \makebox[1.2cm][c]{24.54} &
        \makebox[1.2cm][c]{26.30} \\

        \makebox[2.7cm][l]{\textcolor{gray}{CogVideoX-2b}} & 
        \makebox[0.3cm][c]{\textcolor{gray}{100}} &
        \makebox[0.3cm][c]{\textcolor{gray}{200}} &
        \makebox[0.9cm][c]{\textcolor{gray}{80.01}} &
        \makebox[0.9cm][c]{\textcolor{gray}{67.23}} &
        \makebox[0.9cm][c]{\textcolor{gray}{98.60}} &
        \makebox[0.7cm][c]{\textcolor{gray}{89.98}} &
        \makebox[1.3cm][c]{\textcolor{gray}{49.05}} &
        \makebox[0.7cm][c]{\textcolor{gray}{68.60}} &
        \makebox[1.2cm][c]{\textcolor{gray}{24.04}} &
        \makebox[1.2cm][c]{\textcolor{gray}{25.37}} &
        \makebox[1.2cm][c]{\textcolor{gray}{25.68}} \\

        \midrule[0.75pt]

        \makebox[2.7cm][l]{\textbf{ADM}} & 
        \makebox[0.3cm][c]{8} &
        \makebox[0.3cm][c]{8} &
        \makebox[0.9cm][c]{92.94} &
        \makebox[0.9cm][c]{65.89} &
        \makebox[0.9cm][c]{95.80} &
        \makebox[0.7cm][c]{84.97} &
        \makebox[1.3cm][c]{72.92} &
        \makebox[0.7cm][c]{56.06} &
        \makebox[1.2cm][c]{22.63} &
        \makebox[1.2cm][c]{23.64} &
        \makebox[1.2cm][c]{26.17} \\

        \makebox[2.7cm][l]{\textbf{ADM w/ CFG}} & 
        \makebox[0.3cm][c]{8} &
        \makebox[0.3cm][c]{16} &
        \makebox[0.9cm][c]{89.41} &
        \makebox[0.9cm][c]{69.89} &
        \makebox[0.9cm][c]{97.00} &
        \makebox[0.7cm][c]{71.35} &
        \makebox[1.3cm][c]{81.26} &
        \makebox[0.7cm][c]{53.90} &
        \makebox[1.2cm][c]{21.48} &
        \makebox[1.2cm][c]{23.79} &
        \makebox[1.2cm][c]{25.92} \\

        \makebox[2.7cm][l]{\textcolor{gray}{CogVideoX-5b}} & 
        \makebox[0.3cm][c]{\textcolor{gray}{100}} &
        \makebox[0.3cm][c]{\textcolor{gray}{200}} &
        \makebox[0.9cm][c]{\textcolor{gray}{87.64}} &
        \makebox[0.9cm][c]{\textcolor{gray}{67.34}} &
        \makebox[0.9cm][c]{\textcolor{gray}{99.60}} &
        \makebox[0.7cm][c]{\textcolor{gray}{83.93}} &
        \makebox[1.3cm][c]{\textcolor{gray}{68.24}} &
        \makebox[0.7cm][c]{\textcolor{gray}{56.35}} &
        \makebox[1.2cm][c]{\textcolor{gray}{25.16}} &
        \makebox[1.2cm][c]{\textcolor{gray}{25.82}} &
        \makebox[1.2cm][c]{\textcolor{gray}{27.79}} \\

        \bottomrule[1.5pt]
    \end{tabular}
    \caption{
        VBench~\cite{huang2024vbench} detailed results on separate score for each semantic dimension.
    }
    \vspace{-0.2cm}
    \label{tab:vbench_2}
\end{table*}

\begin{figure*}[t]
    \centering
    \includegraphics[width=\linewidth]{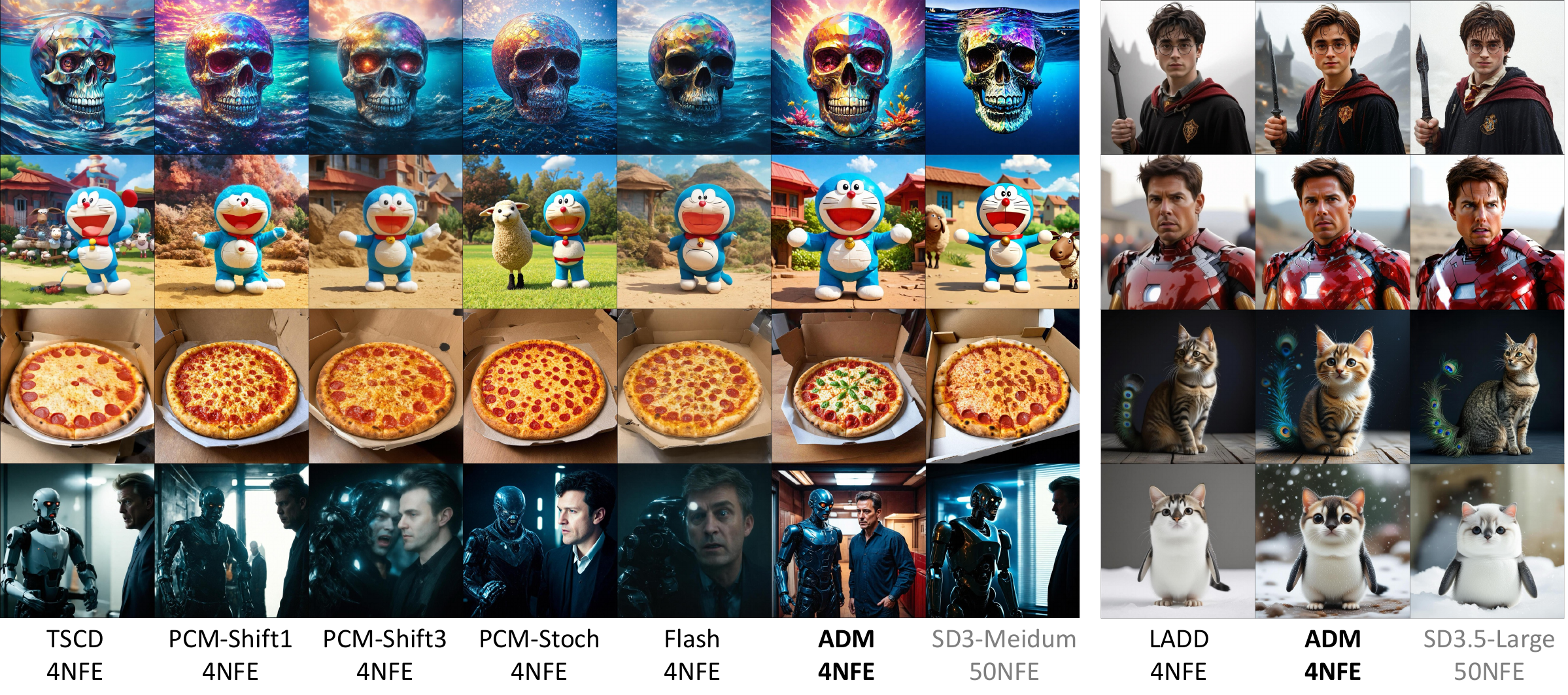}
    \vspace*{-0.6cm}
    \caption{Qualitative results on LoRA fine-tuning SD3-Medium and fully tine-tuning SD3.5-Large.}
    \vspace*{-0.3cm}
    \label{fig:qualitative_sd3}
\end{figure*}






\subsection{Efficient Image Synthesis}\label{appendix:qualitative_fewstep_image}

\cref{fig:qualitative_sd3} qualitatively compares our method with other state-of-the-art distillation techniques on SD3~\cite{esser2024scaling} series models.
The results demonstrate that our method is competitive to the original model in terms of color, detail, structure and image-text alignment, while outperforming other methods including TSCD, PCM~\cite{wang2024phased}, Flash~\cite{chadebec2024flash} and LADD~\cite{sauer2024fast}.

\subsection{Efficient Video Synthesis}\label{appendix:quantitative_comparisons}

\begin{figure*}[t]
    \centering
    \includegraphics[width=0.85\linewidth]{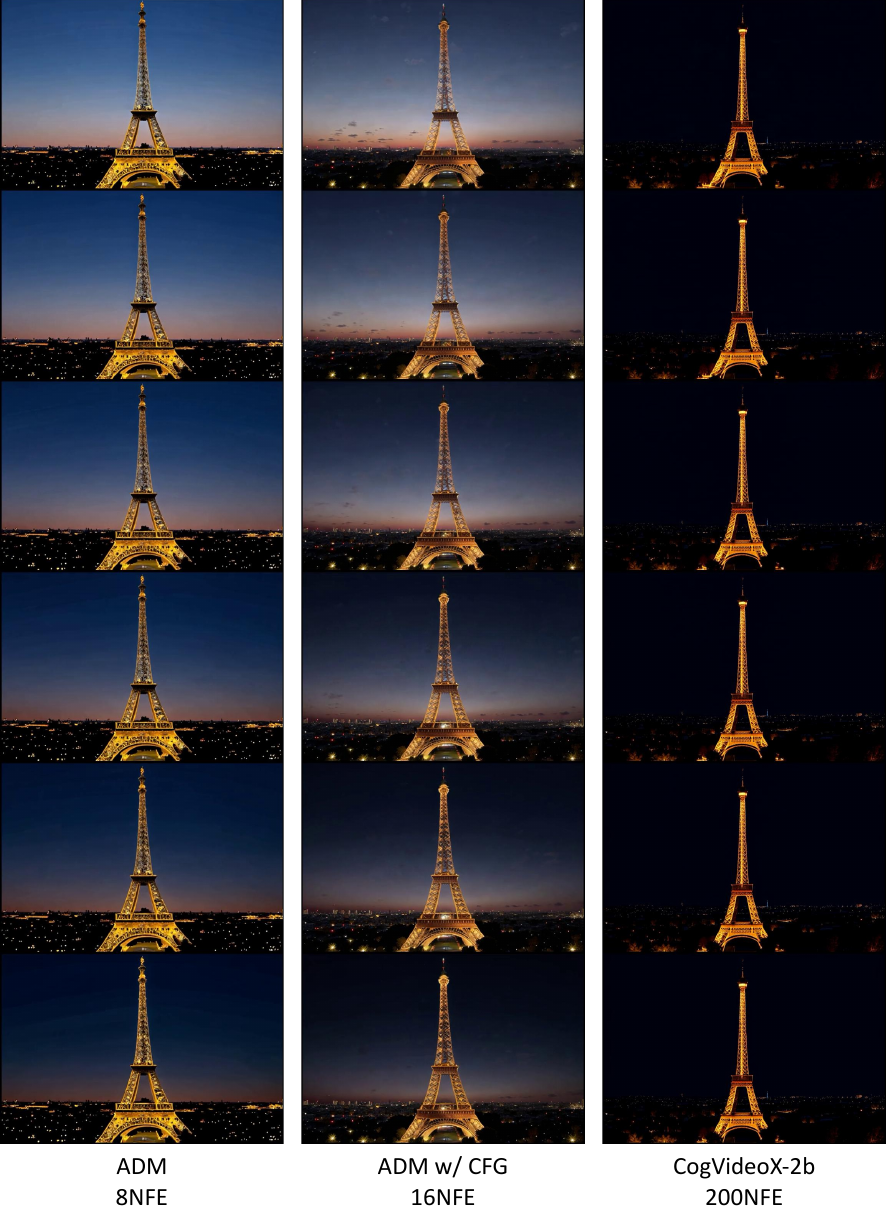}
    \caption{Qualitative comparisons on CogVideoX-2b generators. The random seed has been fixed. \textbf{Prompt:} {A time-lapse sequence captures the transformation of the iconic Eiffel Tower fromdaylight into the evening. The tower, standing tall and majestic in its originalgolden hue, \ul{gradually transitions into a silhouette against the twilight sky}. Asthe sun sets, the city lights begin to flicker on, casting a warm glow over theParisian landscape. The tower's intricate iron lattice structure becomes more defined,its shadow lengthening across the Champ de Mars. The background includes the SeineRiver and the Parisian rooftops, adding depth and context to the scene. As darknessfalls, the Eiffel Tower is illuminated by its own lights, turning into a beaconof Paris, shimmering against the starry backdrop.}}
    \label{fig:qualitative_2b_1}
\end{figure*}

\begin{figure*}[t]
    \centering
    \includegraphics[width=0.85\linewidth]{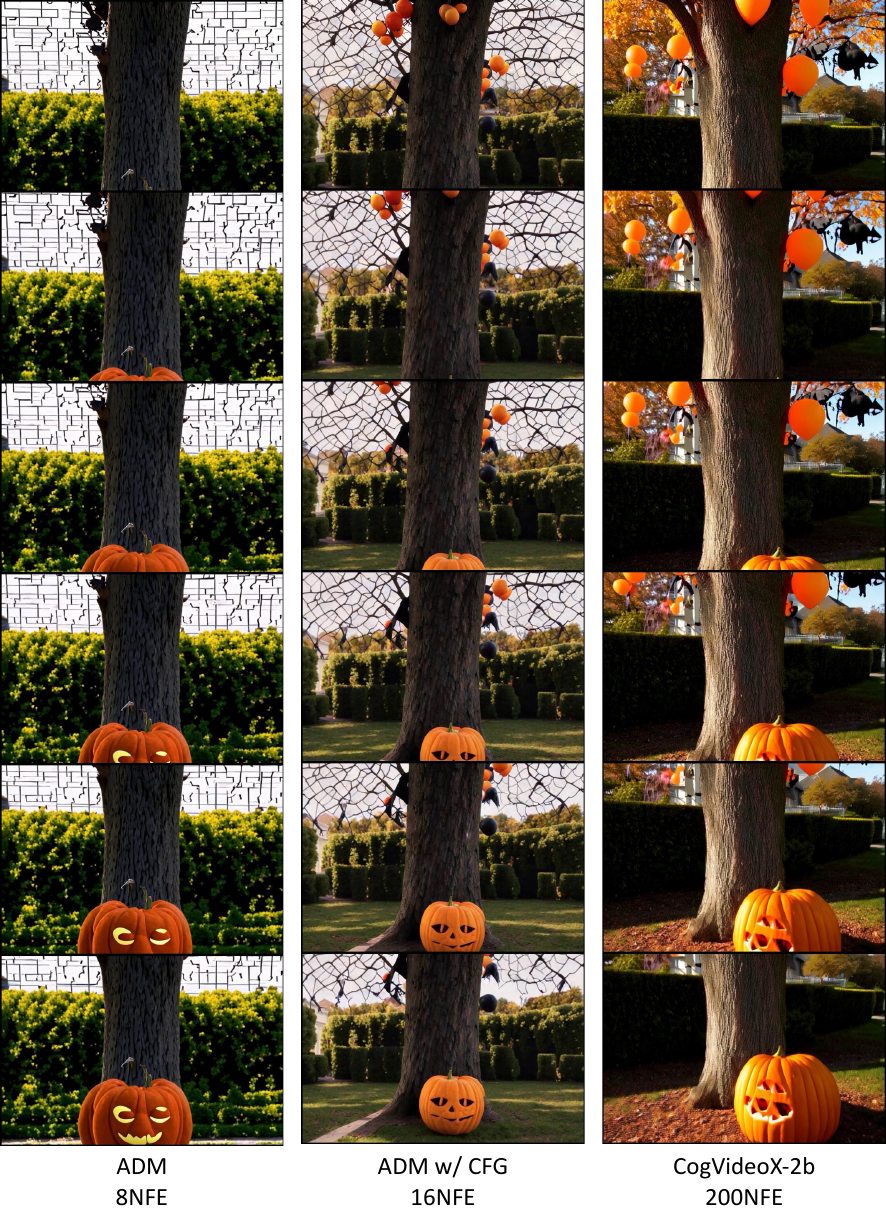}
    \caption{Qualitative comparisons on CogVideoX-2b generators. The random seed has been fixed. \textbf{Prompt:} {A vibrant oak tree, adorned with festive Halloween decorations, stands tall in asuburban backyard. The trunk is thick and sturdy, supporting a variety of decorations.Hanging from its branches are luminous orange and black balloons, spooky spiderwebs,and fluttering ghosts. A \ul{large, carved pumpkin sits at the base}, its intricate faceaglow with a warm, welcoming light. The scene is set against a backdrop of neatlytrimmed hedges and a path leading up to a quaint house, all bathed in the soft glowof autumn sunlight.}}
    \label{fig:qualitative_2b_2}
\end{figure*}

\begin{figure*}[t]
    \centering
    \includegraphics[width=0.85\linewidth]{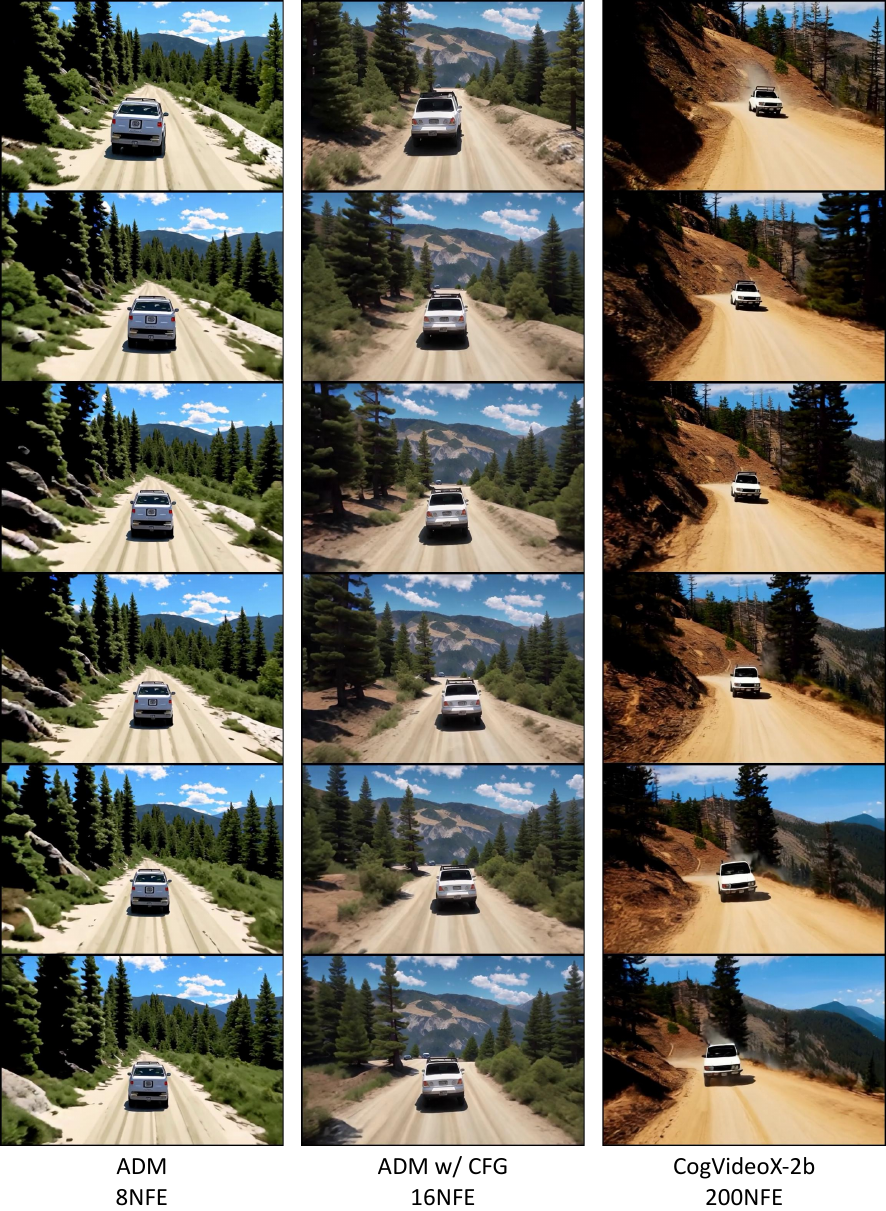}
    \caption{Qualitative comparisons on CogVideoX-2b generators. The random seed has been fixed. \textbf{Prompt:} {The \ul{camera follows behind a white vintage SUV} with a black roof rack as it speedsup a steep dirt road surrounded by pine trees on a steep mountain slope, dust kicksup from it's tires, the sunlight shines on the SUV as it speeds along the dirt road,casting a warm glow over the scene. The dirt road curves gently into the distance,with no other cars or vehicles in sight. The trees on either side of the road areredwoods, with patches of greenery scattered throughout. The car is seen from therear following the curve with ease, making it seem as if it is on a rugged drivethrough the rugged terrain. The dirt road itself is surrounded by steep hills andmountains, with a clear blue sky above with wispy clouds.}}
    \label{fig:qualitative_2b_3}
\end{figure*}

\begin{figure*}[t]
    \centering
    \includegraphics[width=0.85\linewidth]{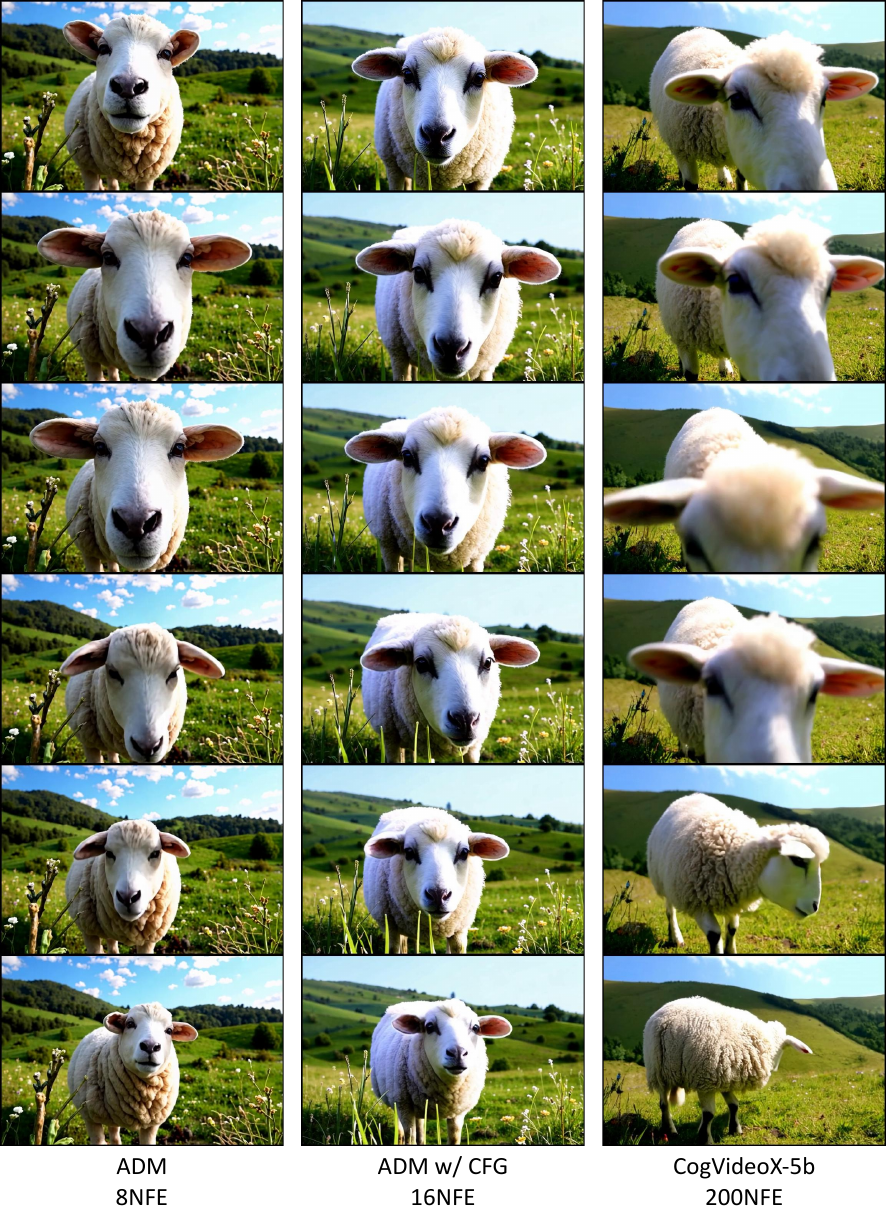}
    \caption{Qualitative comparisons on CogVideoX-5b generators. The random seed has been fixed. \textbf{Prompt:} {A fluffy, white sheep stands in a lush, green meadow, its wool glistening under the warm afternoon sun. The scene transitions to \ul{a close-up of the sheep's gentle face}, its big, curious eyes and soft, twitching ears capturing attention. The background features rolling hills dotted with wildflowers and a clear blue sky. The sheep then grazes peacefully, its movements slow and deliberate, as a gentle breeze rustles the grass. Finally, \ul{the sheep looks up, framed by the picturesque landscape}, embodying tranquility and the simple beauty of nature.
}}
    \label{fig:qualitative_5b_1}
\end{figure*}

\begin{figure*}[t]
    \centering
    \includegraphics[width=0.85\linewidth]{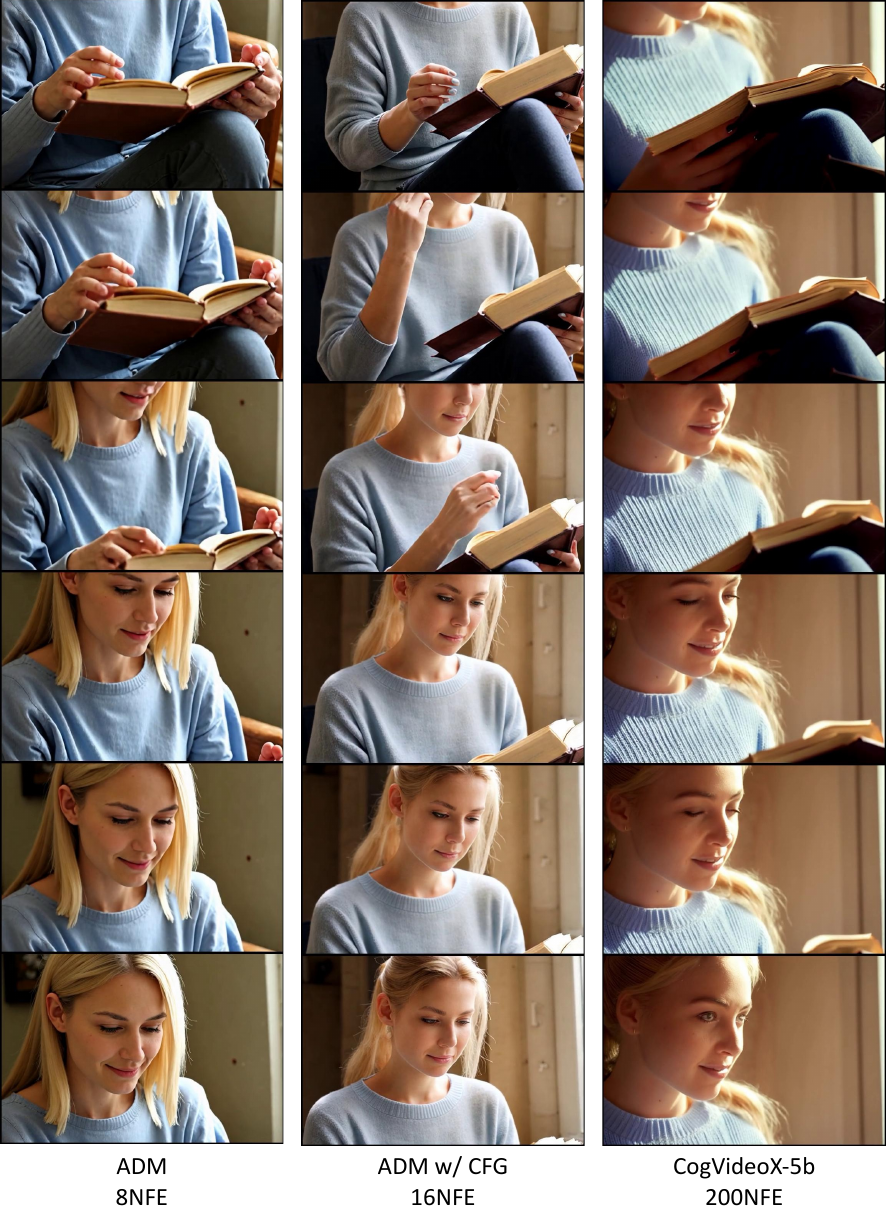}
    \caption{Qualitative comparisons on CogVideoX-5b generators. The random seed has been fixed. \textbf{Prompt:} {Gwen Stacy, with her signature blonde hair tied back in a ponytail, sits in a cozy, sunlit room, engrossed in a thick, leather-bound book. She wears a casual yet stylish outfit: a light blue sweater, dark jeans, and black ankle boots. The camera starts at her hands, delicately turning a page, revealing her neatly painted nails. As \ul{the camera tilts up, it captures her focused expression}, her eyes scanning the text with curiosity and intensity. The warm sunlight filters through a nearby window, casting a soft glow on her face, highlighting her serene and studious demeanor. The scene \ul{ends with a close-up of her thoughtful smile}, suggesting a moment of discovery or reflection.}}
    \label{fig:qualitative_5b_2}
\end{figure*}

\begin{figure*}[t]
    \centering
    \includegraphics[width=0.85\linewidth]{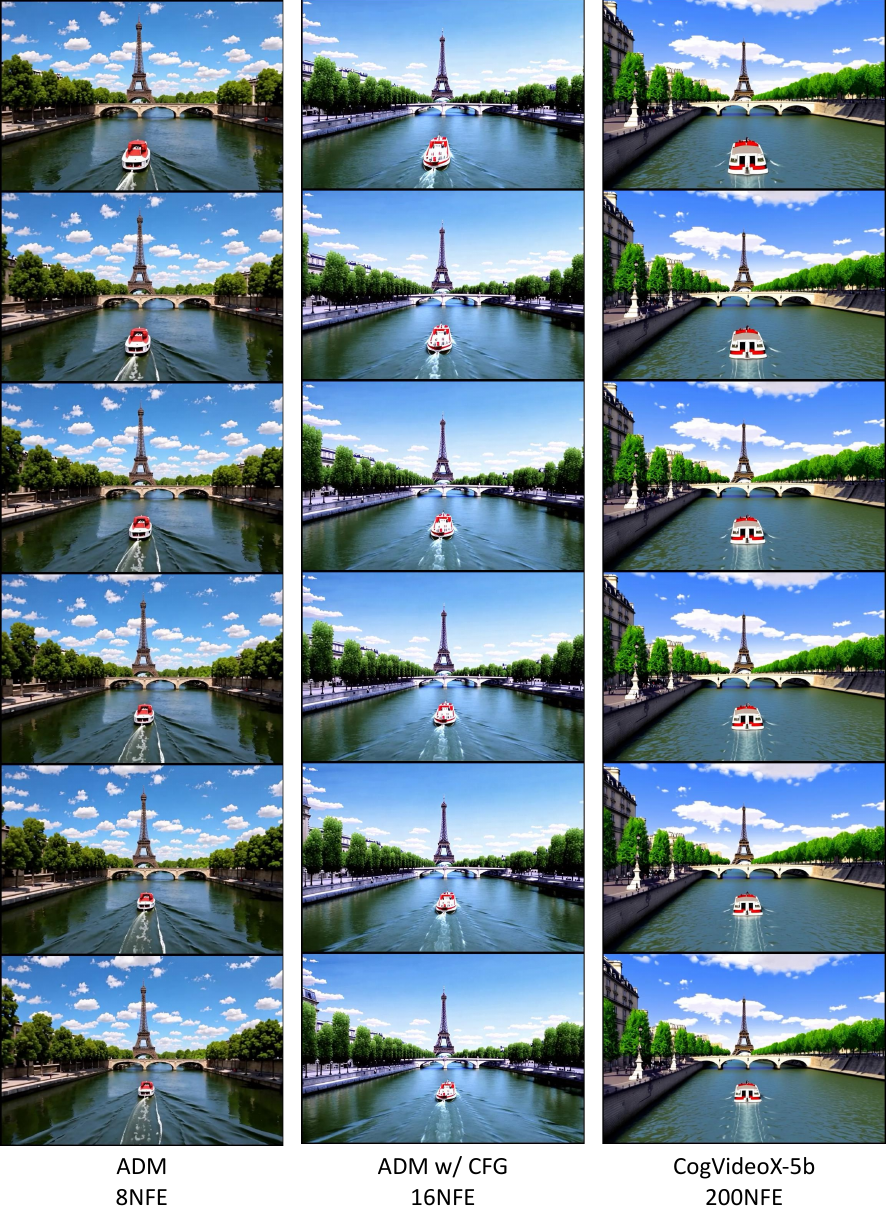}
    \caption{Qualitative comparisons on CogVideoX-5b generators. The random seed has been fixed. \textbf{Prompt:} {A charming boat with a red and white hull sails leisurely along the serene Seine River, its gentle wake creating ripples in the water. The iconic Eiffel Tower stands majestically in the background, framed by a clear blue sky and fluffy white clouds. As \ul{the camera zooms out, the scene expands to reveal lush green trees lining the riverbanks}, quaint Parisian buildings with their classic architecture, and pedestrians strolling along the cobblestone pathways. The boat continues its tranquil journey, passing under elegant stone bridges adorned with ornate lampposts, capturing the essence of a peaceful day in Paris.
}}
    \label{fig:qualitative_5b_3}
\end{figure*}

\cref{tab:vbench_1,tab:vbench_2} present the details of VBench~\cite{huang2024vbench} results on the base model and few-step generators of CogVideoX~\cite{yang2024cogvideox}.
In \cref{fig:qualitative_2b_1,fig:qualitative_2b_2,fig:qualitative_2b_3,fig:qualitative_5b_1,fig:qualitative_5b_2,fig:qualitative_5b_3}, we present several cases for qualitative comparisons between our CogVideoX~\cite{yang2024cogvideox} generators and baseline model.
The results show that our 8-step generators are generally semantically comparable to the original model, even with semantic enhancements on some cases, e.g., the change of light in \cref{fig:qualitative_2b_1} and the movement of the sheep in \cref{fig:qualitative_5b_1}.
While in terms of imaging quality, \ul{generators with CFG are generally more detailed and have more delicate textures} than those without CFG.
The deficiencies in detail are reflected in, for example, the slightly rough hand and the incorrect number of fingers in \cref{fig:qualitative_5b_2}, whereas the one with CFG is much more natural.
As well as the \ul{generator without CFG is also much higher in color contrast}, which visually looks sometimes too vibrant to be sufficiently realistic.
These demonstrate the importance of CFG for text-to-video models, which might not be fully reflected by quantitative metrics.

\begin{figure*}[t]
    \centering
    \includegraphics[width=0.77\linewidth]{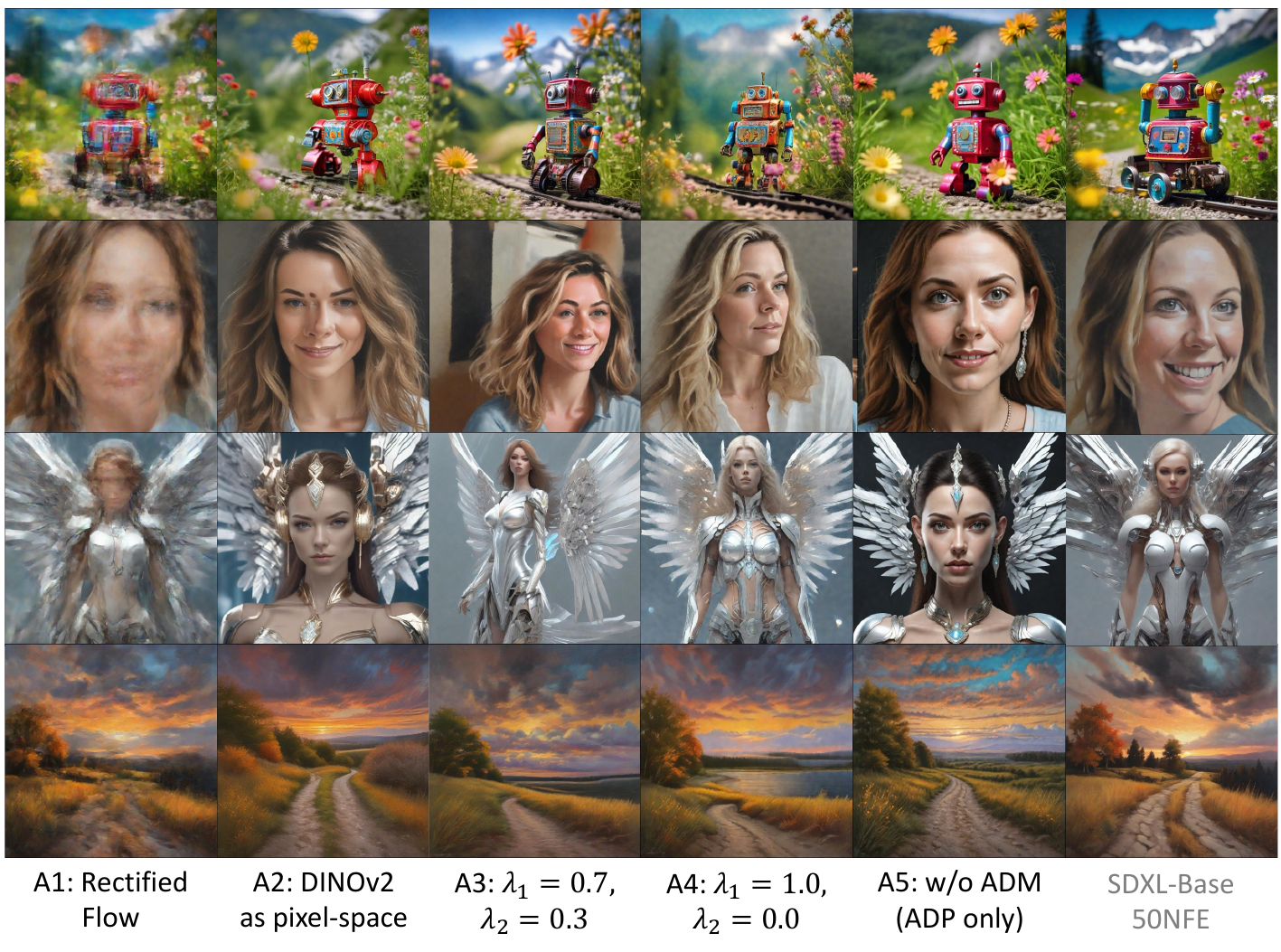}
    \caption{Qualitative comparisons for ablation studies on adversarial distillation.}
    \label{fig:qualitative_ablation_reflow}
\end{figure*}

\begin{figure*}[t]
    \centering
    \includegraphics[width=0.77\linewidth]{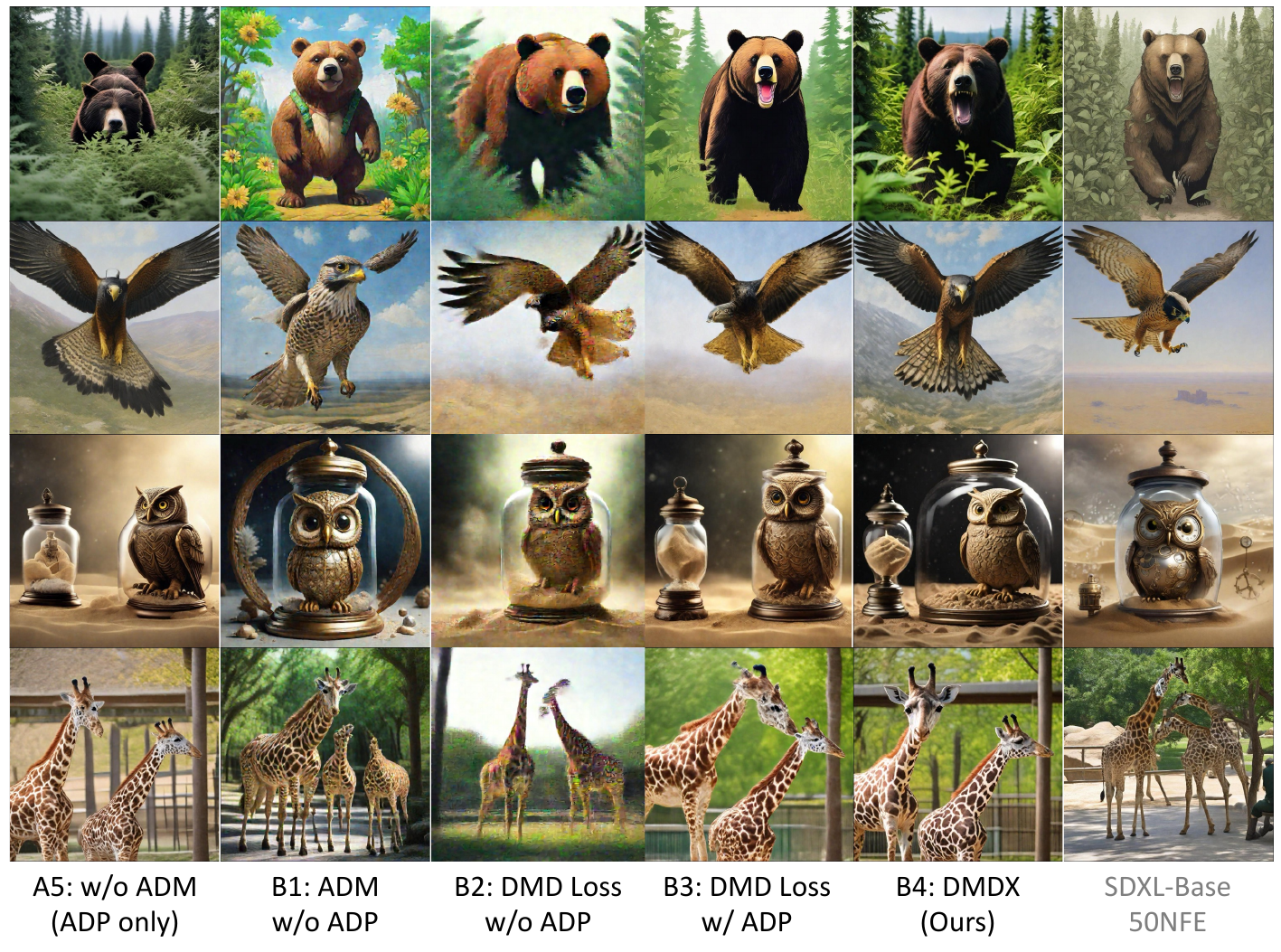}
    \caption{Qualitative comparisons for ablation studies on score distillation.}
    \label{fig:qualitative_ablation_dmdx}
\end{figure*}

\begin{figure*}[t]
    \centering
    \includegraphics[width=\linewidth]{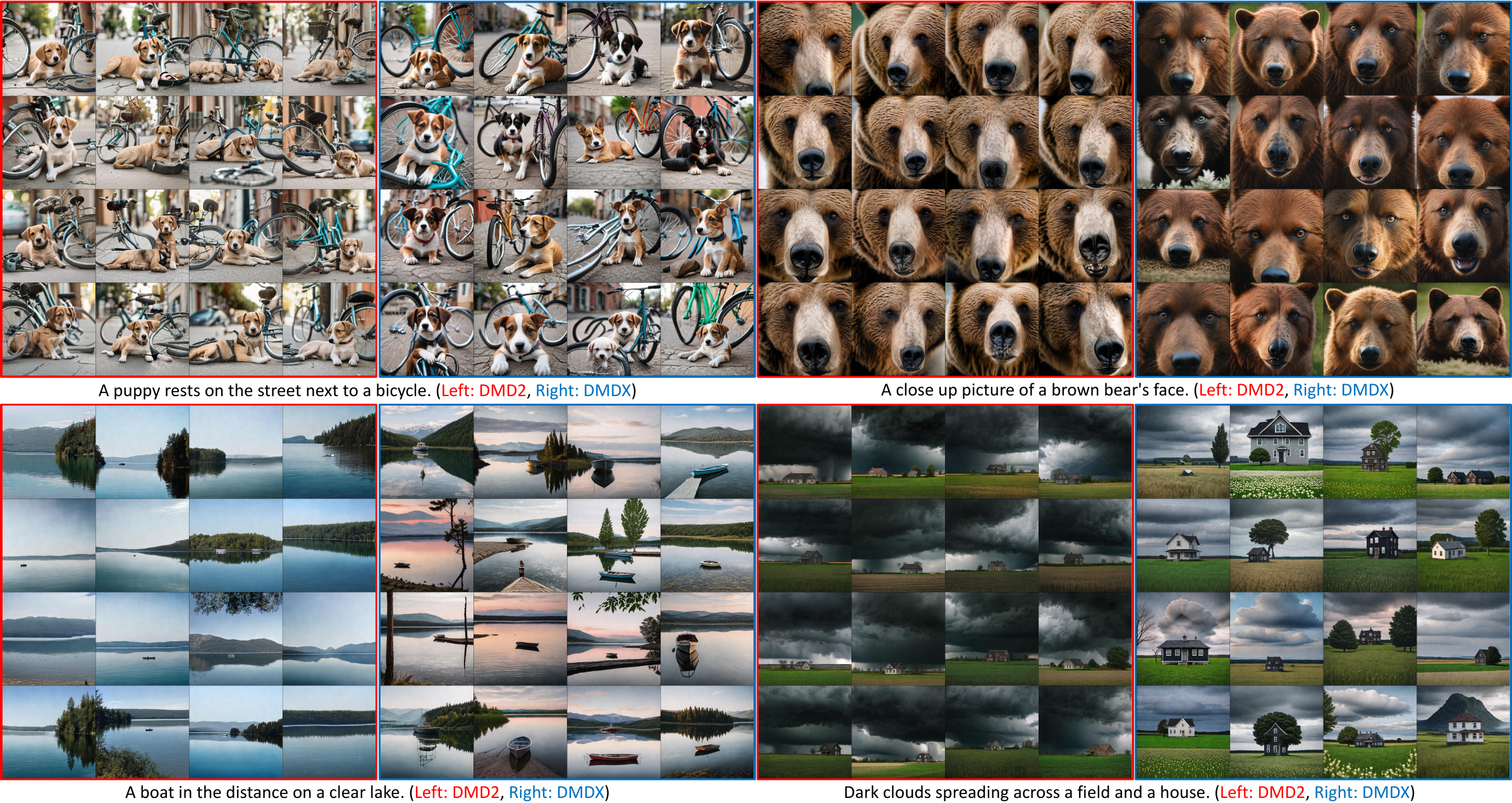}
    \vspace{-0.6cm}
    \caption{
        Qualitative diversity comparisons with DMD2.
    }
    \vspace{-0.2cm}
    \label{fig:diversity}
\end{figure*}

\subsection{Ablation Studies}\label{appendix:qualitative_ablation_study}
As for ablation on adversarial distillation shown in \cref{fig:qualitative_ablation_reflow}, the two main problems with other baseline settings are structure and blurriness.
When using MSE loss for a single reflow process as in Rectified Flow~\cite{liu2022flow}, it is obvious that it is struggling to generate a structurally visible image.
And switching the SAM~\cite{kirillov2023segment} model to DINOv2~\cite{oquab2024dinov2}, we can clearly see the structural collapse of both the robot and the face in the figure, which is unexpected and may be caused by the fact that its input resolution is only 518px, and the images we generate are all 1024px need to be resized before they can be input.
Another possible explanation is that the prior knowledge used by SAM for instance segmentation is richer than that provided by DINOv2 for discriminative self-supervised learning, which facilitates the generation of local fine-grained details.
The structural problems encountered when increasing the weight of pixel-space $\lambda_2$ are similar, while decreasing its weight causes a very noticeable blurring that is clearly visible in the figure, so we suggest setting $\lambda_1=0.85,\lambda_2=0.15$ is a reasonable configuration.

In \cref{fig:qualitative_ablation_dmdx}, we provide qualitative comparisons for ablation studies on score distillation.
Compared to the baseline without ADM (ADP only), we can see that the ADM distillation indeed serves as a fine-tuning process to refine the generator in terms of both color, detail and the most notable structure.
Although standalone ADM can also produce efficient generator, the noise artifact within 1-step generations as similarly observed by \cite{lin2024sdxllightning,yin2024improved} still exists, and with our ADP this issue can be addressed well.
Notably, the visualization results demonstrate that employing the DMD loss without ADP integration induces substantially severe noise artifacts.
Compared to using ADM alone, its qualitative disadvantage is much more pronounced than the gap observed in the quantitative results.
With ADP, the DMD loss generates relatively good results, yet it remains inferior to ADM in terms of visual fidelity and structural integrity.
This indicates that its distribution matching capability is weaker than that of ADM, which is consistent with our analysis in the quantitative results of \cref{sec:ablation}.

Additionally, we showcase additional randomly curated multi-seed samples in \cref{fig:diversity} compared with DMD2, clearly demonstrating that our images exhibit richer variations in texture, color, brightness, contrast and structural composition.

\section{Broader Impact}
Considering that many current methods leverage generated data from foundation models as assistance~\cite{mi2025data}, our acceleration approach for diffusion models can substantially expedite this process, thereby benefiting numerous downstream tasks such as recognition~\cite{xiao2017building}, detection~\cite{ma2025finegrainedzeroshotobjectdetection}, retrieval~\cite{ma2025msdetreffectivevideomoment,lu2022improving}, domain adaptation~\cite{lu2024mlnet,shen2023collaborative}, etc. 
Alternatively, we can train LoRA to acquire an acceleration plugin, enhancing the efficiency of customized vertical models for image~\cite{lu2024coarsefine} or video~\cite{xu2025hunyuanportrait} generation.

\section{Prompt List}\label{appendix:prompt_list}

Below we list the text prompts used for the generated content shown in this paper (from top to bottom, from left to right).
Note that since models like SDXL-Base~\cite{rombach2022highresolution} only use CLIP~\cite{radford2021learning} as a text encoder, which only supports a maximum of 77 tokens, the response and text-image alignment may be insufficient for some long prompts and its limited capacity in understanding.


\textbf{\underline{We use the following prompts for \cref{fig:qualitative_sdxl}:}}
\begin{itemize}
    \item A beautiful woman facing to the camera, smiling confidently, colorful long hair, diamond necklace, deep red lip, medium shot, highly detailed, realistic, masterpiece.
    \item An owl perches quietly on a twisted branch deep within an ancient forest. Its sharp yellow eyes are keen and watchful.
    \item A young badger delicately sniffing a yellow rose, with a lion lurking in the background.
    \item A pickup truck going up a mountain switchback.
\end{itemize}

\textbf{\underline{We use the following prompts for \cref{fig:qualitative_sd3}:}}
\begin{itemize}
    \item A photograph of a giant diamond skull in the ocean, featuring vibrant colors and detailed textures.
    \item A still of Doraemon from "Shaun the Sheep" by Aardman Animation.
    \item A pizza is displayed inside a pizza box.
    \item movie still of a man and a robot in a moment of horror, movie still, cinematic composition, cinematic light, by edgar wright and david lynch
    \item harry potter as a skyrim character
    \item film still of Tom Cruise as Ironman in the Avengers
    \item A beautiful award winning picture of a cute cat in front of a dark background. The cat is a cat-peacock hybrid and has a peacock tail and short peacock feathers on the body. fluffy, extremely detailed, stunning, high quality, atmospheric lighting
    \item a cute animal that's a penguin cat hybrid
\end{itemize}

\textbf{\underline{We use the following prompts for \cref{fig:qualitative_ablation_reflow}:}}
\begin{itemize}
    \item A colorful tin toy robot runs a steam engine on a path near a beautiful flower meadow in the Swiss Alps with a mountain panorama in the background, captured in a long shot with motion blur and depth of field.
    \item A portrait painting of Leighann Vail.
    \item A photo of a mechanical angel woman with crystal wings, in the sci-fi style of Stefan Kostic, created by Stanley Lau and Artgerm.
    \item A painting depicting a foothpath at Indian summer with an epic evening sky at sunset and low thunder clouds.
\end{itemize}

\textbf{\underline{We use the following prompts for \cref{fig:qualitative_ablation_dmdx}:}}
\begin{itemize}
    \item A bear walks through a group of bushes with a plant in its mouth.
    \item A falcon in flight, depicted in a highly detailed painting by Ilya Repin, Phil Hale, and Kent Williams.
    \item A steampunk pocketwatch owl is trapped inside a glass jar buried in sand, surrounded by an hourglass and swirling mist.
    \item Some giraffes are walking around the zoo exhibit.
\end{itemize}

\end{document}